\documentclass[twocolumn]{article} 
\usepackage{arxiv}

\usepackage[utf8]{inputenc} % allow utf-8 input
\usepackage[T1]{fontenc}    % use 8-bit T1 fonts
\usepackage{hyperref}       % hyperlinks
\usepackage{url}            % simple URL typesetting
\usepackage{booktabs}       % professional-quality tables
\usepackage{amsfonts}       % blackboard math symbols
\usepackage{nicefrac}       % compact symbols for 1/2, etc.
\usepackage{microtype}      % microtypography
\usepackage{lipsum}
\usepackage{graphicx}
\graphicspath{ {./images/} }
\usepackage{amssymb}
\usepackage{multirow}
\usepackage{caption} 	 			
\usepackage{booktabs} 		
\usepackage{textcomp}
\usepackage{amsmath,amsfonts}
\usepackage{algorithmic}
\usepackage{algorithm}
\usepackage{array}
\usepackage{hhline}
\usepackage{fancyhdr}
\usepackage{cite}
\usepackage{stfloats}
\usepackage{tabularx}
\usepackage{fancyhdr}
\usepackage{xcolor}
\title{ADA-DPM: A Neural Descriptors-based Adaptive Noise Filtering Strategy for SLAM}

\newif\ifuniqueAffiliation
\uniqueAffiliationtrue

\author{
	\hspace{0.5mm} 
	Yongxin Shao \\ 
	College of Metrology Measurement and Instrument\\
	China Jiliang University\\
	Hangzhou, China 031100 \\
	\texttt{shaoyxpaper@163.com} \\
	\And
	\hspace{0.5mm} 
	Aihong Tan \\
	College of Mechanical and Electrical Engineering\\
	China Jiliang University\\
	Hangzhou, China 031100 \\
	\texttt{Tanah@cjlu.edu.cn} \\
	\And
	\hspace{0.5mm} 
	Binrui Wang \thanks{Binrui Wang is the Corresponding Author.)---\emph{not} for acknowledging funding agencies.} \\
	College of Mechanical and Electrical Engineering\\
	China Jiliang University\\
	Hangzhou, China 031100 \\
	\texttt{wangbrpaper@163.com} \\
	\And
	\hspace{0.5mm} 
	Yinlian Jin \\
	College of Mechanical and Electrical Engineering\\
	China Jiliang University\\
	Hangzhou, China 031100 \\
	\texttt{jinyinglian@cjlu.edu.cn} \\
	\And
	\hspace{0.5mm} 
	Licong Guan \\
	College of Mechanical and Electrical Engineering\\
	China Jiliang University\\
	Hangzhou, China 031100 \\
	\texttt{lcguan941@cjlu.edu.cn} \\
	\And
	\hspace{0.5mm} 
	Peng Liao \\
	College of Mechanical and Electrical Engineering\\
	China Jiliang University\\
	Hangzhou, China 031100 \\
	\texttt{liaopeng202310@163.com} \\
}
\begin{document}
\twocolumn[{
	\maketitle
	\begin{abstract}
		Lidar SLAM plays a significant role in mobile robot navigation and high-definition map construction. However, existing methods often face a trade-off between localization accuracy and system robustness in scenarios with a high proportion of dynamic objects, point cloud distortion, and unstructured environments. To address this issue, we propose a neural descriptors-based adaptive noise filtering strategy for SLAM, named ADA-DPM, which improves the performance of localization and mapping tasks through three key technical innovations. Firstly, to tackle dynamic object interference, we design the Dynamic Segmentation Head to predict and filter out dynamic feature points, eliminating the ego-motion interference caused by dynamic objects. Secondly, to mitigate the impact of noise and unstructured feature points, we propose the Global Importance Scoring Head that adaptively selects high-contribution feature points while suppressing the influence of noise and unstructured feature points. Moreover, we introduce the Cross-Layer Graph Convolution Module (GLI-GCN) to construct multi-scale neighborhood graphs, fusing local structural information across different scales and improving the discriminative power of overlapping features. Finally, experimental validations on multiple public datasets confirm the effectiveness of ADA-DPM.
	\end{abstract}
	\vspace{1.5em}
}]
%\let\thefootnote\relax
%\footnotetext{\small\itshape%
%	This work has been submitted to the IEEE for possible publication. Copyright may be transferred without notice, after which this version may no longer be accessible.}

\section{Introduction}
Autonomous mobile robots are playing an increasingly important role in modern society, and they are of great significance for enhancing work efficiency, addressing challenges confronting humanity, and driving technological innovation\cite{wang2023computed,10509820,SHAO2024106623}. SLAM (Simultaneous Localization and Mapping) is a fundamental technology in robotic environment perception. Its objective is to enable the robot, when moving in an unknown environment without prior knowledge, to use its on-board sensors to construct the spatial map in real time and estimate its own pose\cite{10606294}.

According to the main types of sensors used, mainstream SLAM algorithms can be divided into visual SLAM and LiDAR SLAM. Among them, visual SLAM usually relies on camera sensors. Benefiting from the advantages of cameras such as light weight, simple installation, and low cost, visual SLAM has been widely used in unmanned aerial vehicles (UAVs) and mobile robot platforms with limited load capacity\cite{ZHOU2025126235,LIU2026129799,ZHAO2025128763}. However, the location accuracy of visual SLAM decreases significantly in situations with sudden changes in lighting conditions and lack of scene texture features. Compared with cameras\cite{guan2024dynamic,guan2023instance,mur2017orb,qin2018vins,9351614}, the point cloud collected by LiDAR is less susceptible to environmental lighting conditions and can more effectively preserve the 3D structural information of the environment, making LiDAR SLAM one of the hot research directions attracting numerous scholars currently\cite{CVISIC2022104189,LI2025126487}.

Currently, according to different feature extraction methods, LiDAR SLAM can be mainly divided into handcrafted feature-based methods and deep learning-based methods. The former\cite{zhang2014loam,wang2020intensity,kim2021scan,wang2021f} relies on predefined geometric rules (such as corner features, plane features) to extract features, which easily leads to the problem of sparse or redundant features: in weak-texture scenes (such as tunnels, grasslands, etc.), geometric features are often sparse, which can easily cause the loss of key constraints and result in drift in pose estimation; while in scenarios with extremely rich geometric features (such as urban streets), although the features are dense, it is accompanied by huge computational overhead and high memory usage, making real-time reconstruction of large-scale scenes difficult\cite{zhang2024deeppointmap}. In contrast, deep learning-based methods\cite{Wang_2019_ICCV, xu2021omnet} use neural networks to autonomously extract and fuse geometric features in the scene, enabling more compact and expressive feature representations, thus completing robot localization and mapping tasks more efficiently.

However, in some complex scenarios (such as scenarios lacking fixed landmarks like buildings and roads; scenes with many dynamic objects), deep learning-based methods still face challenges. (1) In unstructured environments (such as rural and semi-urban scenes), point cloud often contains a large amount of noise and outliers, which significantly affect the extraction of local features, cause incorrect matching of feature points\cite{ruchti2018mapping}. What's more, structurally feature points with regular distribution (such as those of buildings and road signs) are far easier to achieve correct matching than feature points with irregular distribution (for example, feature points of vegetation). Ignoring such differences will restrict the accuracy and efficiency of matching. (2) Although deep learning-based methods are generally robust to a small number of dynamic objects (such as pedestrians and vehicles), when the proportion of dynamic objects in the scene is too high, their ego-motion will generate additional local constraints. Existing dynamic point elimination methods (such as introducing auxiliary networks\cite{pfreundschuh2021dynamic}) are effective, but inevitably increase the computational cost of the system. (3) For the loop closure detection task in SLAM, using a single receptive field for feature extraction fails to fully utilize multi-scale neighborhood structural information. As a result, when the receptive field increases, the similarity of point clouds in certain overlapping regions may decrease significantly, which increases the probability of incorrect predictions for repeated scenes and ultimately leads to errors in pose estimation. In general, when existing methods deal with complex scenarios such as dynamic points, noise points, and unstructured environments, they often need to introduce additional constraints. This affects the matching of feature points to a certain extent, and they frequently face a trade-off between location accuracy and system robustness.

To address the above challenges, we propose a neural descriptors-based adaptive noise filtering strategy for SLAM, named ADA-DPM. This method consists of three key modules: the Dynamic Segmentation Head, the Importance Scoring Head, and the GLI-GCN (Cross Layer Intra-Graph Convolution) module.

To adaptively select feature points with higher contributions and suppress the influence of noise and unstructured feature points, the ADA-DPM introduces the Global Importance Scoring module. It assigns an importance score weight to each successfully matched feature point pair, emphasizing the influence of those reliable feature points that make significant contributions to point cloud registration, while filtering out low-importance points (which may be noise or outliers), thereby improving the accuracy of feature point matching and suppressing the negative impact of noise and unstructured feature points. Importantly, we propose the importance score loss, which consists of the global reconstruction loss weighted by importance score and the score loss, to supervise the prediction of global importance score weights. Firstly, the KAN Linear Layer\cite{liu2024kan} is used to directly predict the global rigid transformation of the point cloud to align feature points and compute the reconstruction error. Then, the predicted importance score weights are used to weight the reconstruction error, reducing the reconstruction error caused by low-importance points by predicting smaller importance score weights. Additionally, to prevent the model from reducing the global reconstruction loss by predicting all importance score weights as 0 during the training phase, we further construct the score loss with all labels set to 1 to force the model to predict non-zero score weights. Moreover, by supervising the prediction of global rigid transformations between feature points, implicit supervision is applied to the feature extraction network, encouraging the network to learn features that contribute more to the point cloud registration task.

To mitigate the interference caused by dynamic objects, we propose the dynamic point prediction module (Dynamic Segmentation Head). It predicts the category of each feature point (whether it belongs to the dynamic point or static point) and directly removed the feature points predicted as dynamic points. This avoids dynamic points introducing incorrect local constraints, thereby eliminating the interference of the ego-motion of dynamic objects on the global pose prediction.

Furthermore, to improve the similarity judgment in loop closure detection and explore multi-scale neighborhood structural information, we design the GLI-GCN module. Through cross-layer feature point sampling, it samples the spatial neighborhood features of deep-layer feature points from their corresponding shallow-layer feature point sets and constructs a local neighborhood graph for each deep-layer feature point. This allows the fusion of neighborhood information from different levels, enabling the perception of richer multi-scale geometric features, thereby enhancing feature representation capabilities and particularly improving similarity judgments in loop closure detection.

In summary, the main contributions of this paper are as follows:
\begin{itemize}
	\item{We propose an adaptive noise filtering strategy. The Dynamic Segmentation Head is designed to predict the category of each feature point, directly removed the feature points predicted as dynamic point, thereby eliminating the interference of their ego-motion on the global pose estimation. The Importance Scoring Head is designed to adaptively select feature points with higher contribution, enhancing the contribution of reliable feature points while suppressing erroneous pairings caused by outliers and unstructured feature points.}
	\item{We propose a multi-scale perception module(GLI-GCN ). Through cross-layer feature sampling and local neighborhood graph construction, it fuses neighborhood structural information from different receptive fields to generate local feature descriptors with multi-scale perception capabilities. This significantly alleviates the problem of reduced similarity in the neighborhood structure of overlapping regions caused by a single fixed receptive field.}
	\item{To verify the advantages of ADA-DPM, we conduct sufficient experimental validations on multiple public datasets. The experimental results show that the ADA-DPM method not only ensures good location accuracy but also has excellent robustness to noise points.}
\end{itemize}
\section{Related Work}
According to the different encoding methods of feature points, mainstream SLAM methods can be divided into methods based on handcrafted feature-based method and deep learning-based methods.

\subsection{Handcrafted Feature-based Methods}
Handcrafted feature-based methods rely on traditional geometric algorithms to extract structured features from point clouds (such as line segments, planes, corner points, etc.) for matching relevant feature points between adjacent frames. The Iterated Closest Points (ICP) method proposed by Besl et al.\cite{besl1992method} laid the foundation for inter-frame matching in SLAM, but it is prone to falling into local optima. To overcome this issue, Censi et al.\cite{censi2008icp} proposed the PLICP method, which modifies the distance between nearest points in ICP to the distance from a point to the line connecting its two nearest points; NICP\cite{serafin2015nicp} uses normal vectors to represent local geometric structures (such as walls, ground) and minimizes the positional distance and normal vector angle between two points through the least square method; Deschaud et al.\cite{deschaud2018imls} proposed the low-drift IMLSICP algorithm, which introduces a specific sampling strategy to select representative points and minimize their distance to the target point cloud; Yang et al.\cite{yang2013go} combined the branch-and-bound method for efficient search in 3D space. Although these methods alleviate the local optimum problem of ICP, in sparse point clouds of large-scale scenes, the number of points available for matching is extremely scarce, which limits the location accuracy. To address this issue, Segal et al. proposed a generalized ICP method based on point-to-plane distance\cite{segal2009generalized}; Zhang Ji et al. further combined this scheme with point-to-edge distance and proposed the famous LOAM\cite{zhang2014loam} framework; subsequent improvements include LeGO-LOAM\cite{shan2018lego} and LOAMlinox\cite{xu2021fast}, etc. Although these handcrafted feature-based methods have achieved good results in texture-rich scenes, they are limited by the design of feature extraction rules, resulting in limited generalization ability. In particular, they perform poorly in weakly textured scenes and scenes with extremely rich geometric features \cite{xu2021fast}. 
\subsection{Deep Learning-Based Methods}
Deep learning-based methods utilize neural networks (such as PointNet\cite{qi2017pointnet}, PointNet++\cite{qi2017pointnet++}, etc.) to automatically learn the global semantic features and contextual relationships of point clouds, thereby completing the matching of relevant feature points between adjacent frames. DCP\cite{Wang_2019_ICCV} efficiently realizes the estimation of rigid body transformation between adjacent frame point clouds by using singular value decomposition-based soft matching alignment; PRNet\cite{wang2019prnet} embeds a key point detection module on the basis of DCP and introduces the Gumbel-softmax operator to improve the soft matching in DCP to hard matching, which enhances the registration accuracy of point clouds with local overlap; RPMNet\cite{yew2020rpm} fuses handcrafted designed feature with features encoded by deep learning to construct robust point feature representations, and introduces a doubly stochastic constraint into the learned soft matching matrix to evaluate the inlier confidence of matching pairs and eliminate outliers, thus effectively suppressing the interference of outliers and enhancing the robust rigid body transformation estimation under local overlap. Furthermore, IDAM\cite{li2020iterative} constructs a distance-aware soft matching matrix by deeply fusing spatial geometric information and deep feature information; RGM\cite{fu2021robust} fuses the node information and edge information of point clouds with the help of graph matching algorithms, and uses Sinkhorn iteration to solve graph matching optimization problem for rigid body transformation estimation.

Subsequently, PointNetLK\cite{aoki2019pointnetlk} uses PointNet\cite{qi2017pointnet} to extract global features and modifies the Lucas-Kanade algorithm to avoid the inherent defect that the PointNet representation cannot perform gradient estimation through convolution. RegTR\cite{yew2022regtr} realizes end-to-end point cloud registration in large-scale scenes by virtue of the powerful feature representation ability of Transformer. OMNet\cite{xu2021omnet} and UTOPIC\cite{chen2022utopic} improve the reliability of predicting overlapping regions in complex scenes by learning overlapping region masks and modeling the uncertainty of overlapping region masks. FINet\cite{xu2022finet} proposes a dual-branch rigid body transformation regression model, which realizes robust transformation estimation by decoupling rotation and translation features at the feature level. DeepMapPoint\cite{zhang2024deeppointmap} extracts highly representative and sparse neural descriptors to realize memory-efficient map representation and accurate multi-scale localization tasks (such as odometry and loop closure), thus constructing a SLAM framework completely based on deep learning.

Early studies usually assumed that SLAM is based on fully static scenes to simplify the problem. To deal with dynamic and semi-dynamic targets in real environments, Ruchti et al. calculated the probability of dynamic targets in each frame through a network and filtered out dynamic targets with high probability \cite{ruchti2018mapping}; RangeNet++\cite{milioto2019rangenet++} converted point clouds into depth maps for semantic segmentation, and then re-mapped the points with semantics; Pfreundschuh\cite{pfreundschuh2021dynamic} et al. directly used a point cloud segmentation network to segment the original point cloud and eliminated dynamic points before the mapping process. Although these methods avoid the interference of dynamic objects to a certain extent, they often need to introduce additional networks for assistance, which increases extra computational overhead.
\begin{figure*}
	\centering
	\includegraphics[width=6.5in]{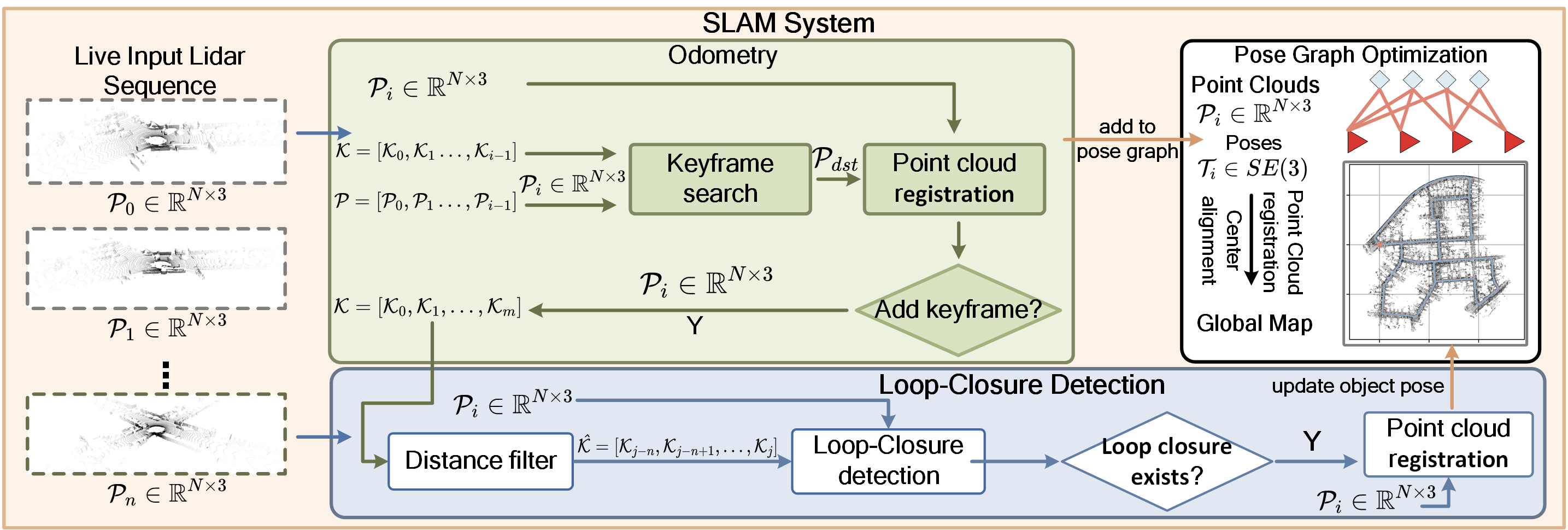}
	\caption{The SLAM pipeline of ADA-DPM. The green part represents the odometry part; the blue part represents loop closure detection; the white part represents pose graph optimization; the blue-boxed portion within the white part represents the final global map and trajectory obtained.}
	\label{fig_1}
\end{figure*}
\section{Methodology}
\subsection{Problem Definition}
The essence of LiDAR SLAM can be attributed to the optimization of dynamic pose estimation: solving for the robot's motion trajectory through frame-to-frame point cloud registration in unknown environments. Its mathematical expression can be formalized as finding the optimal rigid transformation parameters $\mathcal{T} = \{ \mathcal{R}\in SO(3), t\in \mathbb{R}^3 \}$, such that the current frame point cloud and the historical frame point cloud achieve spatial consistency after transformation. 

To facilitate the explanation of our method, we define the source point cloud (current frame point cloud) of the i-th frame as $\mathcal{P}_{src} = \{ x_s \in \mathbb{R}^3 \}_{s=1}^S$, and the target point cloud (typically selected from historical keyframe point clouds) to be registered with it as $\mathcal{P}_{dst} = \{ y_t \in \mathbb{R}^3 \}_{t=1}^T$ (where $x$ and $y$ are the 3D coordinates of points; $S$ and $T$ represent the number of points). We define the subscript ${\cdot_{src}}$ to represent feature descriptors from the source point cloud (i.e., the current frame point cloud), and ${\cdot_{dst}}$ to represent feature descriptors from the target point cloud (i.e., the historical frame point cloud); the symbol $\mathcal{P}$ represents three-dimensional coordinates; the symbol $\mathcal{F}$ represents features.
\subsection{SLAM Pipeline of ADA-DPM}
As shown in Figure \hyperref[fig_1]{1}, SLAM System of ADA-DPM mainly consists of three stages: Odometry, Loop-Closure Detection, and Pose Graph Optimization.

\textbf{Odometry:} The core task of the Odometry stage (as shown in the green part of Figure \hyperref[fig_1]{1}) is to predict the optimal rigid transformation $\mathcal{T}_i$ for the current frame point cloud $\mathcal{P}_i$ (at the $i$-th moment, i.e. $\mathcal{P}_{src}$) relative to the historical frame point cloud set $\mathcal{P}$ (the historical frame point cloud set composed of the first $i-1$ frame point clouds, i.e., historical point clouds), and to determine whether $\mathcal{P}_i$ should be added to the historical keyframe point cloud set $\mathcal{K}$. The specific process is as follows: first, the KNN (K-Nearest Neighbor) is used to search the set $\mathcal{K}$ for the point cloud frame closest $\mathcal{P}_{i-1}$, which is denoted as $\mathcal{P}_{dst}$. Then, the point cloud registration network in ADA-DPM is adopted to predict the $\mathcal{T}_i$ between $\mathcal{P}_{dst}$ and $\mathcal{P}_{i}$. Finally, based on the registration confidence $\tau_i$ (for detailed definition, see Equations \hyperref[eq_1]{(1)} and \hyperref[eq_2]{(2)} in Section 3.C) and RMSE (Root Mean Square Error) between  $\mathcal{P}_{dst}$ and $\mathcal{P}_{i}$, it is determined whether to take $\mathcal{P}_{i}$ as a keyframe point cloud.  In addition, to avoid the computational burden caused by excessively frequent keyframe insertion, $\mathcal{P}_i$ is only inserted into $\mathcal{K}$ when the minimum Euclidean distance of the pose between $\mathcal{P}_i$ and all frames in $\mathcal{K}$ is greater than a certain threshold.

\textbf{Loop-Closure Detection:} The core task of the Loop-Closure Detection stage (as shown in the blue part of Figure \hyperref[fig_1]{1}) is to construct a candidate loop frame set $\hat{\mathcal{K}}$ that may contain loops based on the Euclidean distance of poses between $\mathcal{P}_i$ and $\mathcal{K}$, and use the loop detection component in ADA-DPM to predict whether there exists a loop between $\mathcal{P}_i$ and $\hat{\mathcal{K}}$, so as to correct the global trajectory and map drift caused by accumulated errors. In addition, to avoid frequent loop-closure detection, our method only performs loop-closure detection when the number of keyframes accumulated since the last detection reaches a certain number. To minimize additional computation, keyframes with negligible pose differences are discarded during the candidate loop frame selection process.

\textbf{Pose-Graph Optimization:} 
The Pose-Graph Optimization stage (as shown in the white part of Figure \hyperref[fig_1]{1}) fuses historical observation data and uses global constraints to correct trajectory and map drift. At first, we use the KNN to search the set $\mathcal{P}$ for the $k$ frames closest to $\mathcal{K}_m$. We then calculate the Euclidean distances between these $k$ frames and $\mathcal{P}_i$, discard the point cloud frames that exceed a specific distance threshold, and construct a local map composed of the neighboring frames of $\mathcal{P}_i$. Next, we use the point cloud registration network in ADA-DPM to compute the rigid transformation between $\mathcal{P}_i$ and the local map, and use this transformation to optimize the poses in the pose graph. Finally, we perform graph optimization on the pose graph to obtain the final global map (using the standard pose graph optimization provided in Open3D \cite{zhou2018open3d}).
\subsection{ADA-DPM for Point Cloud Registration}
\begin{figure*}
	\centering
	\includegraphics[width=6.5in]{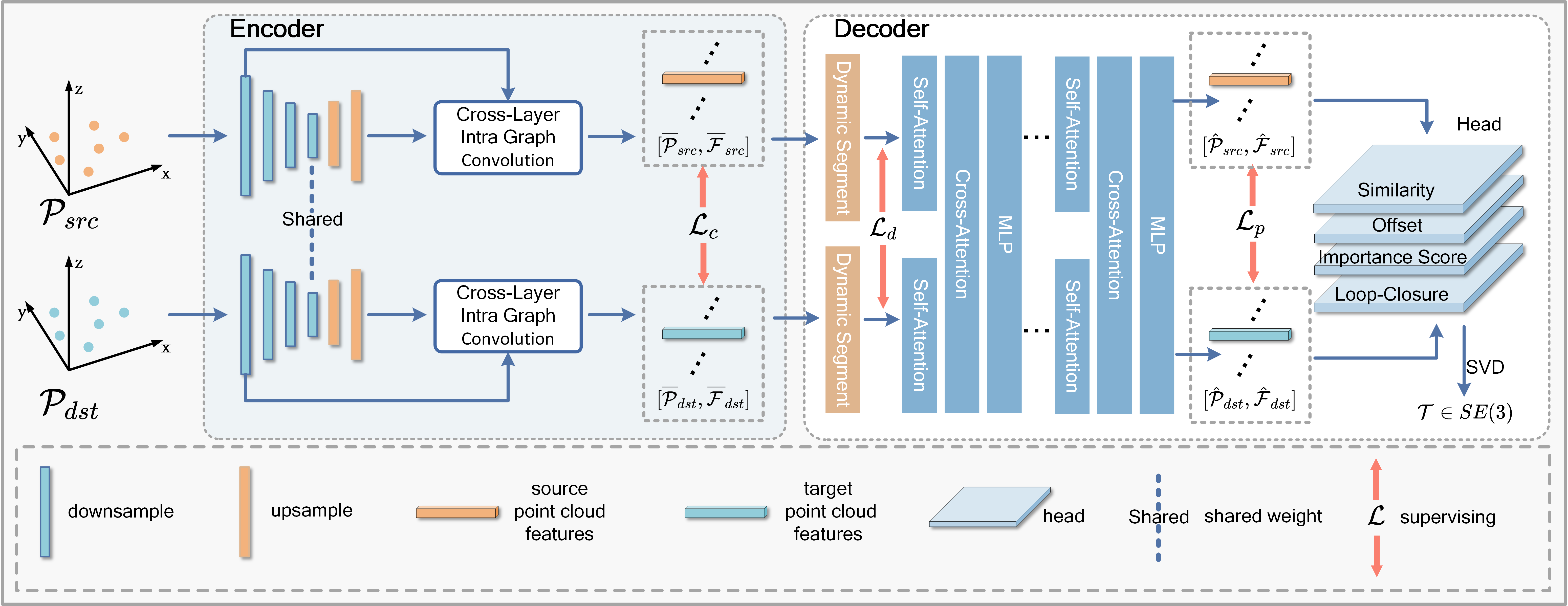}
	\caption{The network overview of ADA-DPM. The blue part represents the Encoder, and the white part represents the Decoder. The networks for both the source point cloud and the target point cloud share the same weights.}
	\label{fig_2}
\end{figure*}
As shown in Figure \hyperref[fig_2]{2}, the ADA-DPM network mainly consists of two parts: Encoder and Decoder. The Encoder performs feature extraction on the input source point cloud $\mathcal{P}_{src}$ and target point cloud $\mathcal{P}_{dst}$, generating the sparse point cloud feature descriptor $[ \overline{\mathcal{P}}, \overline{\mathcal{F}}]$ containing compressed feature information. The Decoder predicts the rigid transformation between the two frames of point clouds and the probability of loop closure based on the feature descriptors $[ \overline{\mathcal{P}}, \overline{\mathcal{F}}]$.

\textbf{Encoder}: As shown in the blue part of Figure \hyperref[fig_2]{2}, the  $\mathcal{P}_{src}$ and $\mathcal{P}_{dst}$ are fed into a weight-sharing feature extraction network, generating the sparse compressed point cloud descriptors for the source point cloud $[ \overline{\mathcal{P}}_{src}, \overline{\mathcal{F}}_{src} ]$ and for the target point cloud $[ \overline{\mathcal{P}}_{dst}, \overline{\mathcal{F}}_{dst} ]$ respectively. The feature extraction network consists of a weight-sharing PointNeXt \cite{qian2022pointnext} and a CLI-GCN module (see Section 3.D for details). PointNeXt downsamples the point cloud and aggregates adjacent geometric features, while the CLI-GCN module aggregates multi-scale local geometric features across receptive fields to enrich the structural information of the point cloud feature.

\textbf{Decoder}: As shown in Figure \hyperref[fig_2]{2}, the Decoder part mainly consists of a feature attention fusion module and multiple decoding heads.

\textbf{Feature Attention Fusion Module}: The feature attention fusion module consists of self-attention module, cross-attention module, and MLP. The self-attention module performs global feature integration on $[ \overline{\mathcal{P}}_{src}, \overline{\mathcal{F}}_{src} ]$ and $[ \overline{\mathcal{P}}_{dst}, \overline{\mathcal{F}}_{dst} ]$ encoded by the Encoder; the cross-attention module exchanges contextual information between the source point cloud and the target point cloud; the MLP integrates the feature information and outputs the decoded compressed point cloud feature descriptors $[ \hat{\mathcal{P}}_{src}, \hat{\mathcal{F}}_{src} ]$ and $[ \hat{\mathcal{P}}_{dst}, \hat{\mathcal{F}}_{dst}]$.

\textbf{Decoding Heads}: The decoding heads in the Decoder mainly include the Similarity Head, Offset Head, Importance Scoring Head, Dynamic Segmentation Head, SVD solver, and Loop-Closure Head. Among them, the point cloud registration process primarily utilizes the Similarity Head, Offset Head, Dynamic Segmentation Head, Importance Scoring Head, and SVD solver, while the Loop-Closure Head is used in Loop-Closure Detection (detailed in Section 3.D). 

In the point cloud registration process, the Dynamic Segmentation Head predicts and filters out dynamic points in $[ \overline{\mathcal{P}}, \overline{\mathcal{F}} ]$ output by the Encoder, before they enter the feature attention module. After the feature attention module, the Similarity Head performs feature point pair matching based on the similarity between $[ \hat{\mathcal{P}}_{src}, \hat{\mathcal{F}}_{src} ]$ and $[ \hat{\mathcal{P}}_{dst}, \hat{\mathcal{F}}_{dst} ]$, and predicts the matching confidence $\tau$ (mathematically described in Equations  \hyperref[eq_1]{(1)}, \hyperref[eq_2]{(2)}). The Offset Head predicts the offset between the compressed feature descriptors of the source point cloud and target point cloud after successful matching($[ \tilde{\mathcal{P}}_{src}, \tilde{\mathcal{F}}_{src} ]$ and $[ \tilde{\mathcal{P}}_{dst}, \tilde{\mathcal{F}}_{dst} ]$), eliminating the misalignment between them in the global coordinate system. Then, the Importance Scoring Head provides an initial estimation of the rigid transformation between $[ \tilde{\mathcal{P}}_{src}, \tilde{\mathcal{F}}_{src} ]$ and $[ \tilde{\mathcal{P}}_{dst}, \tilde{\mathcal{F}}_{dst} ]$, assigns contribution scores to the matched feature point pairs, and discards those with low contribution scores. Finally, the SVD is used to calculate the rigid transformation between the successfully matched point pairs.
\begin{equation}
	\label{eq_1}
	\mathcal{M}_{s} = \left[ (\| \mathcal{F}_{src} \|_{c})^{T} @ (\| \mathcal{F}_{dst} \|_{r}) \right]
\end{equation}
\begin{equation}
	\label{eq_2}
	\tau = \operatorname{TopK}_{r} \left[ \operatorname{Softmax}_{r}(\mathcal{M}_{s}) \odot \operatorname{Softmax}_{c}(\mathcal{M}_{s}) \right]
\end{equation}
\begin{equation}
	\label{eq_3}
	\left[ \tilde{\mathcal{P}}_{src}, \tilde{\mathcal{P}}_{dst} \right] = \left[ \mathcal{P}_{src}, f_{\tau\rightarrow j}(\mathcal{P}_{dst}) \right]
\end{equation}
where $\mathcal{M}_s$ represents the correlation matrix between $[ \hat{\mathcal{P}}_{src}, \hat{\mathcal{F}}_{src} ]$ and $[ \hat{\mathcal{P}}_{dst}, \hat{\mathcal{F}}_{dst} ]$; $\| \cdot \|_r$ and $\| \cdot \|_c$ denote row L2 normalization and column L2 normalization, respectively; $@$ represents matrix multiplication; $\mathcal{F}_{src}$ and $\mathcal{F}_{dst}$ are the features of the source and target point clouds obtained by concatenating $[ \hat{\mathcal{P}}_{src}, \hat{\mathcal{F}}_{src} ]$ and $[ \hat{\mathcal{P}}_{dst}, \hat{\mathcal{F}}_{dst} ]$ along the feature dimension, respectively; $\tau$ denotes the matching confidence between $[ \hat{\mathcal{P}}_{src}, \hat{\mathcal{F}}_{src} ]$ and $[ \hat{\mathcal{P}}_{dst}, \hat{\mathcal{F}}_{dst} ]$; $\operatorname{Softmax}_r(x)$ and $\operatorname{Softmax}_c(x)$ represent Softmax operations applied to rows and columns, respectively; $\operatorname{TopK}_r(x)$ denotes extracting the maximum value from each row; $\odot$ represents element-wise multiplication; $f_{\tau \rightarrow j}( \hat{\mathcal{P}}_{dst} )$ is the mapping function that extracts the target point cloud compressed descriptor corresponding to $\tilde{\mathcal{P}}_{src}$ from $\hat{\mathcal{P}}_{dst}$ based on $\tau$.
\begin{figure}[!t]
	\centering
	\includegraphics[width=3.25in]{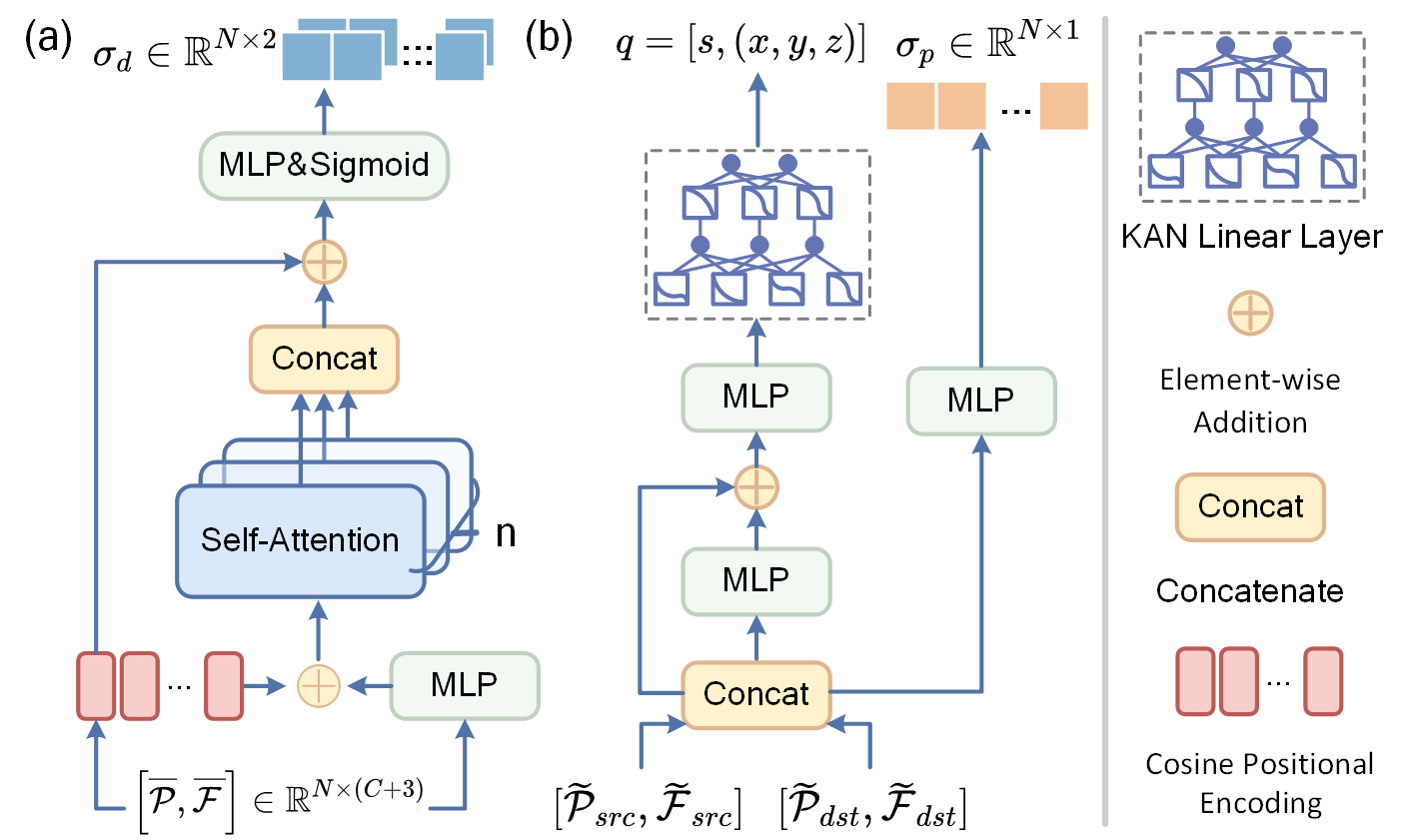}
	\caption{Overview of Dynamic Segmentation Head and Importance Scoring Head. (a) is the overview of Dynamic Segmentation Head; (b) is the overview of Importance Scoring Head.
	}
	\label{fig_3}
\end{figure}

\textbf{Dynamic Segmentation Head}: The constructed Dynamic Segmentation Head consists of a multi-head attention module and an MLP \& Sigmoid classifier, with its structure shown in Figure \hyperref[fig_3]{3(a)}. The Dynamic Segmentation Head deeply mines the global information of $[ \overline{\mathcal{P}}, \overline{\mathcal{F}} ]$ by simply stacking multi-head self-attention modules. After concatenation, the MLP \& Sigmoid is used to predict the probability $\sigma_d \in \mathbb{R}^{N \times 2}$ that $[ \overline{\mathcal{P}}, \overline{\mathcal{F}} ]$ belongs to dynamic points or static points (where $N$ represents the number of feature points in $[ \overline{\mathcal{P}}, \overline{\mathcal{F}} ]$).

In addition, for hard negative samples that are difficult to distinguish due to the lack of semantic information in LiDAR point clouds, we adopt the online hard example mining training strategy\cite{Shrivastava_2016_CVPR}: simultaneously predict the probability $\sigma_d$ of belonging to dynamic points and static points, select $k$ static points with the smallest predicted probability based on the probability of being predicted as static points, and add them to the calculation of the loss function $\mathcal{L}_d$ (to avoid imbalance between positive and negative samples during training, $k$ can be at most 3 times the number of dynamic points). This strategy optimizes the model's ability to distinguish between dynamic and static points.

\textbf{Importance Scoring Head}: The Importance Scoring Head adopts an architecture of stacked MLP and KAN Linear Layer \cite{liu2024kan}, as shown in Figure \hyperref[fig_3]{3(b)}. The Importance Scoring Head treats the preliminary estimation of the rigid transformation between $[ \tilde{\mathcal{P}}_{src}, \tilde{\mathcal{F}}_{src} ]$ and $[ \tilde{\mathcal{P}}_{dst}, \tilde{\mathcal{F}}_{dst} ]$ as a regression task. By decoding $[ \tilde{\mathcal{P}}_{src}, \tilde{\mathcal{F}}_{src} ]$ and $[ \tilde{\mathcal{P}}_{dst}, \tilde{\mathcal{F}}_{dst} ]$, it predicts the rigid transformation between them and scores the matching situation as $\sigma_p \in \mathbb{R}^{N \times 1}$. 
Recently, Liu et al. \cite{liu2024kan} proposed a learnable KAN based on the Kolmogorov-Arnold theory. Unlike traditional approaches that rely on discontinuous activation functions, it learns the combination coefficients of continuous functions rather than the weights of node values, demonstrating strong performance in solving mathematical models. Therefore, during decoding, considering that the rigid transformation between the source and target point clouds satisfies strict mathematical constraints, we use the KAN Linear Layer as the decoder (its mathematical representation\cite{wang2025s2kan} is shown Equations \hyperref[eq_4]{(4)} and \hyperref[eq_5]{(5)}).
\begin{equation}
	\label{eq_4}
	\Phi(x_i) = 
	\begin{bmatrix}
		\phi_{i,1,1}(\cdot) & \phi_{i,1,2}(\cdot) & \cdots & \phi_{i,1,j}(\cdot) \\
		\phi_{i,2,1}(\cdot) & \phi_{i,2,2}(\cdot) & \cdots & \phi_{i,2,j}(\cdot) \\
		\vdots & \vdots & \ddots & \vdots \\
		\phi_{i,k,1}(\cdot) & \phi_{i,k,2}(\cdot) & \cdots & \phi_{i,k,j}(\cdot)
	\end{bmatrix}
\end{equation}
\begin{equation}
	\label{eq_5}
	\mathcal{KAN}(x) = 
	(\Phi_{n-1} \cdot \Phi_{n-2} \cdots \Phi_1 \cdot \Phi_0)(x)
\end{equation}
where $\Phi(\cdot)$ is the activation function matrix of the KAN layer; $\phi_{(i,j,k)}(\cdot)$ represents the learnable continuous activation function connecting the $k$-th neuron in the $i$-th layer and the $j$-th neuron in the $(i+1)$-th layer; $\mathcal{KAN}(\cdot)$ denotes a KAN Linear Layer.

It should be noted that directly predicting the rigid transformation $\mathcal{T} \in SE(3)$ between $\tilde{\mathcal{P}}_{src}$ and $ \tilde{\mathcal{P}}_{dst}$ requires 12 variables to be predicted. To reduce the optimization difficulty, we adopt a simplification strategy: only predict the rotation transformation, which is represented using the quaternion $q = [s, (x, y, z)]$ (where $s$ denotes the real part of $q$, and $x, y, z$ are the imaginary parts of $q$); the translation transformation can be calculated using Equation \hyperref[eq_6]{(6)}. Specifically once $q$ is successfully predicted, we use $q$ to align the $\mathcal{P}_{src}$ and $\mathcal{P}_{dst}$. At this point, the difference in the centroid coordinates of the two aligned sets corresponds to the translation vector. In addition, by supervising $q$, implicit supervision can be imposed on the feature extraction network, enabling it to learn to extract feature points that contribute more to the point cloud registration task.
\begin{equation}
	\label{eq_6}
	\tilde{t} = \frac{q}{\|q\|} \tilde{\mathcal{P}}_{src}^{ctr} \left( \frac{q}{\|q\|} \right)^{-1} - \tilde{\mathcal{P}}_{dst}^{ctr}
\end{equation}
where $(\cdot)^{-1}$ denotes the inverse of $q$; $\|q\|$ represents the norm of $q$; $\tilde{\mathcal{P}}_{src}^{ctr}$, $\tilde{\mathcal{P}}_{dst}^{ctr}$ represent the center point coordinates of $\mathcal{P}_{src}$ and $\mathcal{P}_{dst}$, respectively.

To ensure that the Importance Scoring Head can properly score the matching of feature points, during the training phase, we specifically design the importance scoring loss $\mathcal{L}_s$ to supervise the prediction of $q$ and $\sigma_p$. $\mathcal{L}_s$ consists of the global reconstruction loss weighted by importance score  $\mathcal{L}_q$ and the scoring loss $\mathcal{L}_\sigma$, where $\mathcal{L}_q$ is used to supervise the prediction of $q$, and $\mathcal{L}_\sigma$ is used to supervise the prediction of $\sigma_p$. Specifically, $\mathcal{L}_q$ is represented by the global reconstruction loss weighted by $\sigma_p$; $\mathcal{L}_\sigma$ is represented by cross-entropy with labels all being 1. Since the network weights are optimized in the direction that reduces $\mathcal{L}_s$ during training, for correctly paired point pairs, $\mathcal{L}_q$ can be reduced by predicting the correct $q$; for incorrectly paired point pairs, $\mathcal{L}_q$ needs to be reduced by predicting a smaller $\sigma_p$. Meanwhile, to prevent the model from reducing $\mathcal{L}_q$ by predicting small (or even all-zero) $\sigma_p$ for all point pairs, $\mathcal{L}_\sigma$ calculates the loss with a vector of all ones. This ensures that for correctly paired point pairs, the model predicts $\sigma_p$ as close to 1 as possible, and $\mathcal{L}_q$ can only be reduced by predicting the correct global rigid transformation, thus avoiding the trivial solution where $\sigma_p$ is extremely small or all zero in this case. At this point, $\sigma_p$ can be used to represent the contribution of point pairs to registration.
\subsection{ADA-DPM for Loop-Closure Detection}
The structure of ADA-DPM for Loop-Closure Detection also consists of the Encoder and Decoder, as shown in Figure \hyperref[fig_2]{2}. Among them, the feature attention fusion modules in the Encoder and Decoder adopt the same structure as the corresponding parts in ADA-DPM for Point Cloud Registration and share weights. The Loop-Closure Head follows the same strategy as DeepPointMap\cite{zhang2024deeppointmap}: it predicts the probability of a loop closure between the source point cloud and the target point cloud by stacking two layers of MLPs with shared weights.

For the Encoder part, we adopt the structure of PointNeXt \cite{qian2022pointnext} combined with CLI-GCN. Existing point cloud feature extraction methods (such as PointNet++ \cite{qi2017pointnet++}, VoxelNet \cite{zhou2018voxelnet}, and PointNeXt \cite{qian2022pointnext} used in this paper) aggregate local features through downsampling. As a result, although the finally extracted compressed feature descriptors have rich semantic features, they suffer from issues such as a single receptive field and low feature resolution. For points in overlapping regions, changes in the size of their neighborhood structures do not affect their similarity; however, for points at the edges of overlapping regions, an increase in the receptive field will cause a sharp decrease in the similarity of their neighborhood structures, which may lead to incorrect loop-closure predictions and introduce errors into robot localization and map construction. To make up for this defect of PointNeXt, we construct the CLI-GCN module, which constructs neighborhood graph structures across feature layers to capture geometric features at different scales and emphasizes key geometric features, thereby enhancing the representation capability of the geometric features.

\begin{figure}[!t]
	\centering
	\includegraphics[width=3.25in]{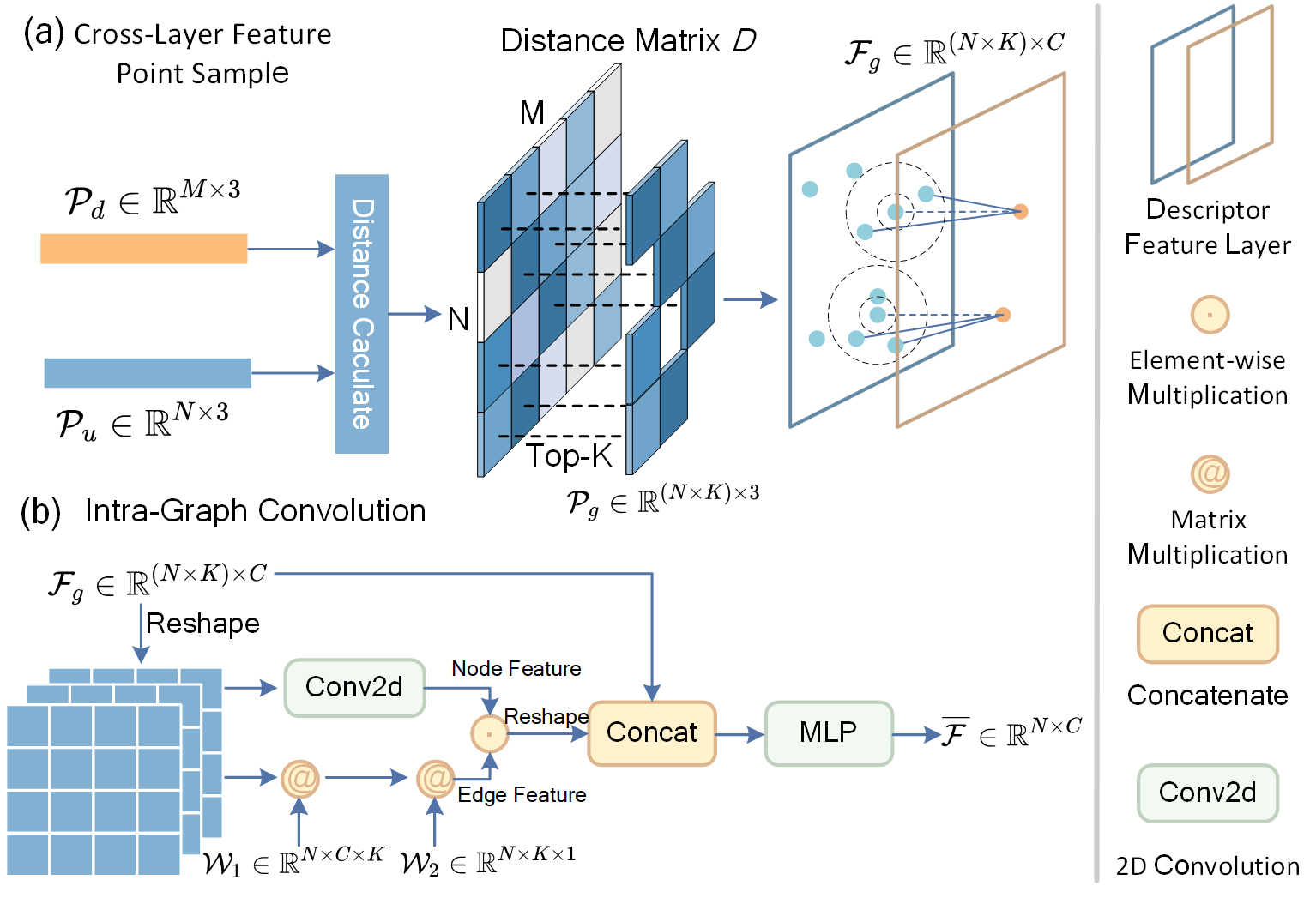}
	\caption{Overview of CLI-GCN. (a) is the overview of Cross-Layer Feature Point Sample; (b) is the the overview of Intra-Graph Convolution.
	}
	\label{fig_4}
\end{figure}

The CLI-GCN structure is shown in Figure \hyperref[fig_4]{4}, mainly consisting of two parts: (a) Cross-Layer Feature Point Sample: this part searches for neighborhood points for deep feature points within the shallow feature point set (performing neighborhood sampling based on distance), constructing multi-scale neighborhood correlations between deep feature points and shallow associated feature points to enhance the expressive ability of geometric features; (b) Intra-Graph Convolution: based on the sampling results, the local neighborhood graph is constructed and feature integration is performed to fuse cross-layer sampled multi-scale features, generating $[ \overline{\mathcal{P}}, \overline{\mathcal{F}}]$.

\textbf{Cross-Layer Feature Point Sample}: The Cross-Layer Feature Point Sample structure is shown in Figure \hyperref[fig_4]{4(a)}. For a given point cloud $\mathcal{P}$, feature extraction is first performed using PointNeXt to obtain shallow features $\mathcal{P}_d$ which are densely distributed but have a small receptive field, and deep features $\mathcal{P}_u$ which are sparsely distributed but have a large receptive field. In the figure, $\mathcal{P}_d \in \mathbb{R}^{M \times 3}$ represents the 3D coordinates of the shallow features, i.e., the features of the first downsampling layer in PointNeXt; $\mathcal{P}_u \in \mathbb{R}^{N \times 3}$ represents the 3D coordinates of the deep features, i.e., the features of the last upsampling layer in PointNeXt (where $M$ and $N$ denote the number of feature points in $\mathcal{P}_u$ and $\mathcal{P}_d$ respectively). Then, the distance between each point in $\mathcal{P}_d$ and each point in $\mathcal{P}_u$ is calculated, resulting in an $M \times N$ distance matrix $\mathcal{D}$. For each row in the distance matrix, the $K$ feature points with the smallest values are selected (with $K=2$ as an example in Figure \hyperref[fig_4]{4(a)}), i.e., for each deep feature point, the $K$ nearest shallow feature points $\mathcal{F}_g \in \mathbb{R}^{(N \times K) \times C}$ (where $C$ denotes the feature dimension) are selected.

\begin{figure}[!t]
	\centering
	\includegraphics[width=3.25in]{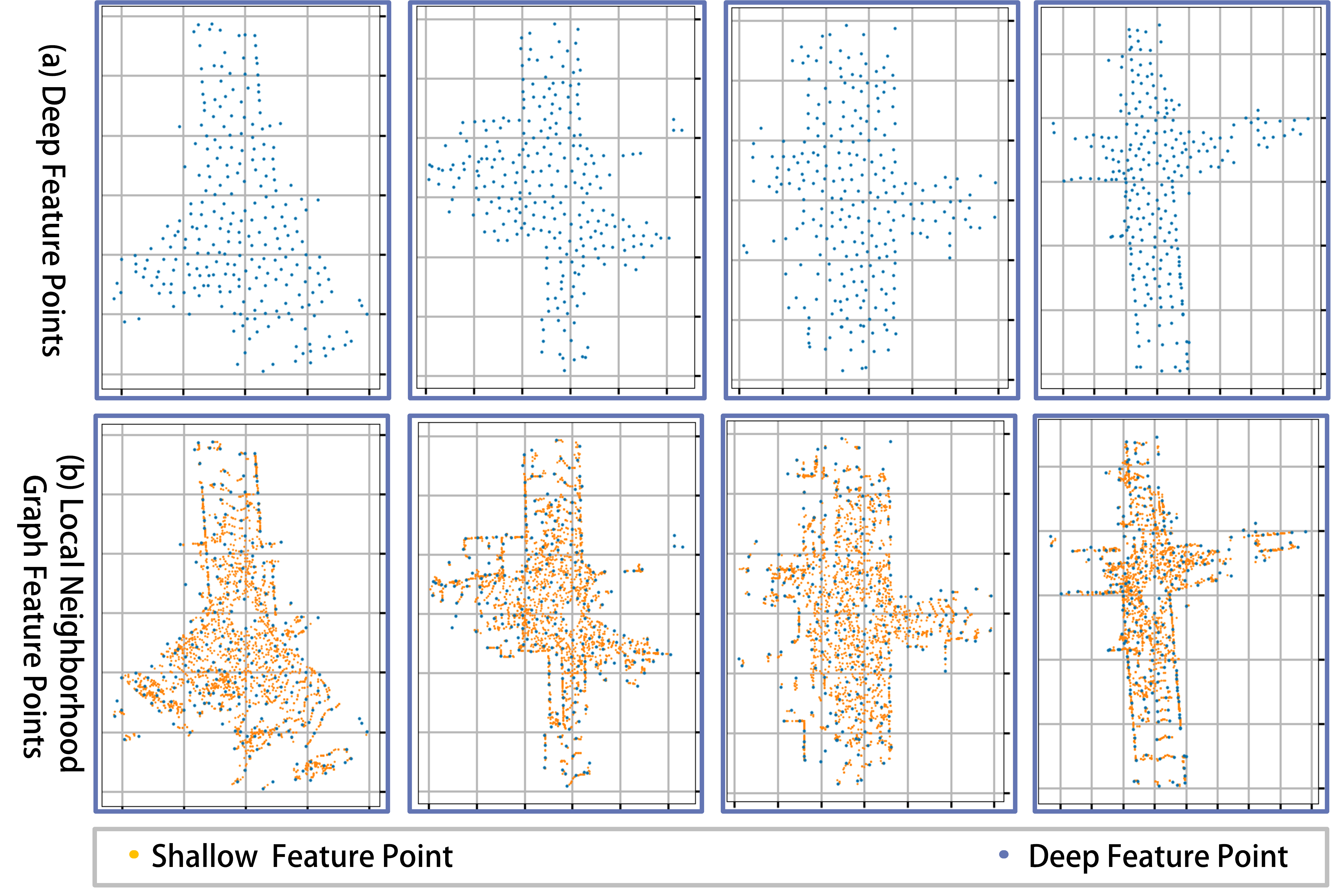}
	\caption{Visualization results of deep feature points and sampled local neighborhood graph feature points. The yellow point denotes the shallow feature point; the blue point denotes the deep feature point.}
	\label{fig_5}
\end{figure}

In addition, to avoid selecting feature points that overlap with deep feature points in 3D space or are extremely close to them, we set upper and lower distance thresholds ($0.5m$ and $1.0m$ in this paper), prioritizing the selection of shallow feature points within the distance threshold range. When the number of shallow feature points within the threshold range is less than $K$, supplementation is performed from feature points with distances greater than the upper threshold. Figure \hyperref[fig_5]{5} shows the visualization results of deep feature points and feature points of the sampled local neighborhood graph. In Figure \hyperref[fig_5]{5}, blue points represent deep feature points, and orange points represent sampled shallow feature points. It can be seen that the blue feature points express more global structures, while the feature points of the local neighborhood graph obtained after sampling retain more local details, containing multi-scale geometric features of the point cloud.

\textbf{Intra-Graph Convolution:} The structure of the Intra-Graph Convolution is shown in Figure \hyperref[fig_4]{4(b)}. It constructs a local neighborhood graph using the cross-layer sampled features $\mathcal{F}_g \in \mathbb{R}^{(N \times K) \times C}$ output by the Cross-Layer Feature Point Sample and performs feature integration to generate $[ \overline{\mathcal{P}}, \overline{\mathcal{F}} ]$. It consists of two branches: the edge feature branch and the node feature branch, which are used to generate edge features and node features of the local neighborhood graph, respectively.

In the node feature branch, $\mathcal{F}_g \in \mathbb{R}^{(N \times K) \times C}$ is first reconstructed into a 2D matrix $\mathcal{F}^{'}_{g} \in \mathbb{R}^{N \times K \times C}$. After reconstruction, each row of $\mathcal{F}^{'}_{g}$ corresponds to the neighborhood of a deep feature point, and the features within each row represent the shallow features in that neighborhood. Then, 2D convolution is used to extract features and generate node features $\mathcal{F}_n$. It should be noted that we use a 2D convolution with a kernel size of $1 \times 3$ for feature extraction, which specifically aggregates features only within the same row. This design ensures that the neighborhoods of each deep feature point remain mutually independent, thereby avoiding mutual interference between node features across different neighborhoods.

In the edge feature branch, the reconstructed $\mathcal{F}^{'}_g \in \mathbb{R}^{N \times K \times C}$ is matrix-multiplied with two learnable weight matrices $\mathcal{W}_1 \in \mathbb{R}^{N \times C \times K}$ and $\mathcal{W}_2 \in \mathbb{R}^{N \times K \times 1}$ to obtain edge features $\mathcal{F}_e \in \mathbb{R}^{N \times K \times 1}$, which represent the correlation between the shallow feature points in each neighborhood and their corresponding deep feature points. 

Finally, the node features and edge features are fused using element-wise multiplication (highlighting key node features with strong correlations) and concatenated with $\mathcal{F}_g \in \mathbb{R}^{(N \times K) \times C}$, and feature integration is performed through an MLP to generate $\overline{\mathcal{F}} \in \mathbb{R}^{N \times C}$:
\begin{equation}
	\label{eq_7}
	\overline{\mathcal{F}} = MLP \left( Cat\left( Reshape\left( \mathcal{F}_{g}^{\prime} @ \mathcal{W}_{1} @ \mathcal{W}_{2} \right) \odot  \mathcal{F}_{n},\ \mathcal{F}_{g} \right) \right)
\end{equation}
where $MLP(\cdot)$ denotes the Multi-Layer Perceptron; $@$ denotes matrix multiplication; $\mathcal{F}_n$ represents the node features; $Cat(\cdot)$ denotes the Concatenate operation; $Reshape(\cdot)$ denotes the Reshape operation.
\subsection{Loss Function}
The training process is divided into two stages: the point cloud registration training stage and the loop-closure detection training stage. The loss function consists of the following components: Coarse Pairing Loss $\mathcal{L}_c$; Pairing Loss $\mathcal{L}_p$; Offset Loss $\mathcal{L}_o$; Dynamic Segmentation Loss $\mathcal{L}_d$; Importance Scoring Loss $\mathcal{L}_s$; Loop-Closure Detection Loss  $\mathcal{L}_{loop}$.

Given the feature point pairs $[ p_{i}^{src}, f_{i}^{src}] \in [ \hat{\mathcal{P}}_{src}, \hat{\mathcal{F}}_{src} ] $ and $[ p_{j}^{dst}, f_{j}^{dst}] \in [ \hat{\mathcal{P}}_{dst}, \hat{\mathcal{F}}_{dst}]$, classify the paired samples into positive samples $\mathcal{Q}_+$, neutral samples $\mathcal{Q}_o$, and negative samples $\mathcal{Q}_-$ based on the distance between $p_{i}^{src}$ and $p_{j}^{dst}$, , following the rules of \hyperref[eq_8]{(8)} to \hyperref[eq_10]{(10)}.
\begin{multline}
	\label{eq_8}
	\mathcal{Q}_{+} = \left\{ \left( p_{i}^{src}, p_{j}^{dst} \right) \mid 
	\arg \min_{j} \left\| p_{i}^{src} - p_{j}^{dst} \right\|_{2}, \right. \\ 
	\left. \left\| p_{i}^{src} - p_{j}^{dst} \right\|_{2} \leq \varepsilon_{positive} 
	\right\}
\end{multline}
\begin{equation}
	\label{eq_9}
	\mathcal{Q}_{o}=\left\{\left(p_{i}^{src}, p_{j}^{dst}\right) \mid \left\|p_{i}^{src}-p_{j}^{dst}\right\|_{2} \leq \varepsilon_{positive}\right\}
\end{equation}
\begin{equation}
	\label{eq_10}
	\mathcal{Q}_{-}=\left\{\left(p_{i}^{src}, p_{j}^{dst}\right) \mid \left\|p_{i}^{src}-p_{j}^{dst}\right\|_{2} > \varepsilon_{positive}\right\}
\end{equation}
where $\varepsilon_{positive}$ denotes the distance threshold, and this paper sets the hyperparameter $\varepsilon_{positive}=1m$.

\textbf{Coarse Pairing Loss \& Pairing Loss:} We use the InfoNCE Loss \cite{oord2018representation} to supervise $[ \overline{\mathcal{P}}, \overline{\mathcal{F}} ]$ and $[ \hat{\mathcal{P}}, \hat{\mathcal{F}} ]$, ensuring that the features of the source point cloud and the target point cloud are similar:
\begin{equation}
	\label{eq_11}
	\mathcal{L}_{c} = -\log \left( \frac{\sum_{\mathcal{Q}_{+}} \exp \left( \hat{f}_{i}^{src} \odot \hat{f}_{j}^{dst} / \alpha \right)}{\sum_{Q} \exp \left( \hat{f}_{i}^{src} \odot \hat{f}_{j}^{dst} / \alpha \right)} \right) \\
\end{equation}
\begin{equation}
	\label{eq_12}
	\mathcal{L}_{p} = -\log \left( \frac{\sum_{\mathcal{Q}_{+}} \exp \left( \hat{f}_{i}^{src} \odot \hat{f}_{j}^{dst} / \alpha \right)}{\sum_{\mathcal{Q}_{+} \cup \mathcal{Q}_{o}} \exp \left( \hat{f}_{i}^{src} \odot \hat{f}_{j}^{dst} / \alpha \right)} \right)
\end{equation}
where $\mathcal{Q}=\mathcal{Q}_+ \cup \mathcal{Q}_- \cup \mathcal{Q}_o$, and this paper sets hyperparameter $\alpha=0.2$.

\textbf{Offset Loss:}  
We follow the method in DeepPointMap, where the Mahalanobis distance is used to compute the Offset Loss:
\begin{equation}
	\label{eq_13}
	\mathcal{L}_{o}=\frac{1}{\left|\mathcal{Q}_{+} \cup \mathcal{Q}_{o}\right|} \sum_{\mathcal{Q}_{+} \cup \mathcal{Q}_{o}} \left\|\delta_{i,j} - \delta_{i,j}^{*}\right\|_{\Sigma}
\end{equation}
where $ \delta_{i,j} $ represents the offset predicted by the Offset Head; $ \delta_{i,j}^* $ represents the ground-truth offset; $ \| \cdot \|_\Sigma $ denotes the Mahalanobis distance.

\textbf{Importance Scoring Loss:}  $ \mathcal{L}_s $ is divided into $ \mathcal{L}_q $ , which supervises the prediction of $q$, and $ \mathcal{L}_\sigma $ which supervises $ \sigma_p$. Among them, $ \mathcal{L}_\sigma $ is represented by the binary cross-entropy with a ground-truth of all ones. The mathematical representation of $ \mathcal{L}_s $ is as follows:
\begin{equation}
	\label{eq_14}
	\begin{split}
		\mathcal{L}_{q} & = \frac{1}{|\mathcal{Q}_{+} \cup \mathcal{Q}_{o}|} \\
		& \sum_{\mathcal{Q}_{+} \cup \mathcal{Q}_{o}} \sigma^{p}_{i,j} \Big\| \frac{q}{\|q\|} p_{i}^{src} \left( \frac{q}{\|q\|} \right)^{-1} \quad + \tilde{t} - p_{j}^{dst} \Big\|_{2}
	\end{split}
\end{equation}
\begin{equation}
	\label{eq_15}
	\mathcal{L}_{s} = \lambda_{q} \mathcal{L}_{q} + \lambda_{\sigma} \mathcal{L}_{\sigma}
\end{equation}
where,  $\tilde{t}$ denotes the translation component, and its data representation is as shown in Equation \hyperref[eq_7]{(7)}; $\sigma^{p}_{i,j}$ denotes the importance score predicted by the Importance Scoring Head; this paper sets the hyperparameter $\lambda_{q}, \lambda_{\sigma}$ are 0.5 and 0.2 respectively.

\textbf{Dynamic Segmentation Loss \& Loop-Closure Detection Loss:}  We use cross-entropy to train the Dynamic Segmentation Head and the Loop-Closure Head. During the training of the Dynamic Segmentation Head, feature points corresponding to moving objects such as cars, pedestrians, and bicycles (including potentially movable objects like parked cars at road signs) are treated as dynamic points, while the remaining feature points are treated as static points. It is important to note that during the Loop-Closure training stage, we freeze the weights of the Encoder and Decoder components and only train the Loop-Closure Detection Head.

The loss function for the point cloud registration part of ADA-DPM is as follows:
\begin{equation}
	\label{eq_16}
	\mathcal{L}_{r g}=\lambda_{1}\mathcal{L}_{c}+\lambda_{2}\mathcal{L}_{p}+\lambda_{3}\mathcal{L}_{o}+\lambda_{4}\mathcal{L}_{s}+\lambda_{5}\mathcal{L}_{d}
\end{equation}
where this paper sets the hyperparameter $\lambda_{1}$, $\lambda_{2}$, $\lambda_{3}$, $\lambda_{4}$, and $\lambda_{5}$ are 0.1, 1.0, 1.0, 1.0, and 1.0, respectively.
\section{Experiments}
\subsection{Implementation Details}
\textbf{Data Augmentation:} (1) Random Sampling:For each input frame of point cloud data, randomly sample 16,384 points. If the number of input points is less than 16,384, pad with zeros to reach this count. (2) Random Occlusion \cite{zhang2024deeppointmap}: Randomly mask $1–3$ regions (removing all points within these regions) within an angular range of $10^{\circ}–150^{\circ}$ and a distance range of $2m–10m$. (3) Random Rigid Transformation: Apply random rigid transformations to the input point cloud. Specifically, the rotation transformation is sampled from a uniform distribution with the mean of 0 and the standard deviation of $\pi$; the translation transformation is sampled from a uniform distribution with the mean of 0 and the standard deviation of 3. (4) Random Local Map Registration: Randomly sample $0–n $ frames of point clouds based on their distance from the current frame to construct a local map. Considering that in practical SLAM tasks, registration is usually made between the point cloud of the current frame and the local map built from historical frames, which expands over time. During training, we set $n$ to increase gradually as the number of training epochs increases. $n$ starts at 2, doubles every 3 epochs, and has a maximum value of 16.

\textbf{Training:} We use the AdamW optimizer\cite{reddi2019convergence} with an initial learning rate of $ lr = 10^{-3} $, weight decay $ wd = 10^{-4} $, combined with a cosine learning rate scheduler\cite{smith2019super}. The training is conducted on 3 RTX 3090 GPUs. For all tasks, the network is trained for 12 epochs, and the model from the final epoch is used for evaluation. 
\begin{table*}[ht]
	\caption{Localization accuracy on SemanticKITTI odometry benchmark(TRANS. and ROT.$\downarrow$). The \textbf{bold} indicates the best result, and the \underline{underline} indicates the second-best result.}
	\label{table1}
	\centering
	% \LARGE
	\resizebox{\linewidth}{!}{      				
		\begin{tabular}{c c |c c c c   c c c c c  c }  	
			\toprule
			\midrule
			\multirow{2}{*}{Modality} &
			\multirow{2}{*}{Method} &  	
			\multicolumn{2}{c}{06} & 
			\multicolumn{2}{c}{07} &
			\multicolumn{2}{c}{08} &
			\multicolumn{2}{c}{09} &
			\multicolumn{2}{c}{10} \\ 
			\cmidrule(lr){3-4} \cmidrule(lr){5-6} \cmidrule(lr){7-8} \cmidrule(lr){9-10} \cmidrule(lr){11-12} 
			& & Trans. & Rot.& Trans. & Rot.& Trans. & Rot.& Trans. & Rot.& Trans. & Rot.\\ 
			\midrule
			\multirow{8}{*}{LiDAR} &LOAM\cite{zhang2014loam}& 0.65&-&0.63&-&1.12&-&0.77&-&0.79&- \\
			&LO-Net\cite{li2019net}& -&-&0.56&0.45&1.08&0.43&0.77&0.38&0.92&0.41\\
			&ISC-LOAM\cite{wang2020intensity}&0.76&0.41&0.56&0.43&1.20&0.50&1.40&0.59&1.87&0.62\\
			&SC-LeGO-LOAM\cite{kim2021scan}&2.54&1.15&2.48&1.78&2.30&1.24&5.37&2.78&10.50&3.79\\
			&F-LOAM\cite{wang2021f}&0.84&0.33&0.88&0.62&\underline{0.87}&0.33&1.03&0.32&1.20&0.29\\
			&LiODOM\cite{garcia2022liodom}&0.83&0.29&0.88&0.61&\textbf{0.86}&0.33&1.03&0.32&1.20&0.29\\
			&LiLO\cite{vizzo2021poisson}&0.54&0.32&0.60&0.61&1.07&0.41&\underline{0.63}&0.32&0.99&0.33\\
			&DeepPointMap\cite{zhang2024deeppointmap}&0.77&-&\underline{0.44}&-&1.09&-&0.95&-&0.69&-\\
			\midrule
			\multirow{5}{*}{Camera} &VISO2\cite{geiger2011stereoscan}&0.79&0.51&1.46&1.13&1.62&0.66&0.84&0.64&1.29&0.64\\
			&ORB-SLAM2\cite{mur2017orb}&0.89&0.27&0.89&0.50&1.03&0.31&0.86&0.25&0.62&0.29\\
			&Vins-Fusion\cite{qin2018vins}&1.35&0.71&1.21&0.90&1.83&0.72&1.82&0.53&2.64&1.01\\
			&OV$^{2}$-SLAM\cite{9351614}&1.13&0.28&1.03&0.57&1.11&0.31&0.96&\textbf{0.20}&\textbf{0.52}&0.18\\
			&SOFT2\cite{CVISIC2022104189}&0.60&\underline{0.23}&0.45&\underline{0.29}&0.91&\textbf{0.26}&0.75&\underline{0.22}&0.74&\textbf{0.24}\\
			&HVL-SLAM\cite{10606294}&0.54&\textbf{0.22}&0.53&\underline{0.29}&1.02&\underline{0.29}&0.64&0.27&0.70&\underline{0.28}\\
			\midrule
			\multirow{3}{*}{LiDAR\&Camera}&DEMO\cite{zhang2017real}&0.96&-&1.16&-&1.24&-&1.17&-&1.14&-\\
			&DVL-SLAM\cite{shin2020dvl}&0.92&-&1.26&-&1.32&-&0.66&-&0.70&-\\
			&CR-LDSO\cite{yuan2023cr}&\underline{0.51}&-&1.01&-&1.14&-&\textbf{0.56}&-&\underline{0.60}&-\\
			\midrule
			LiDAR&ours&\textbf{0.34}&0.29&\textbf{0.35}&\textbf{0.28}& \textbf{0.86} &0.34&0.79&0.31&\underline{0.60}& \underline{0.28}\\
			\midrule
			\bottomrule
		\end{tabular}
	}
\end{table*}
\begin{figure*}
	\centering
	\includegraphics[width=6.5in]{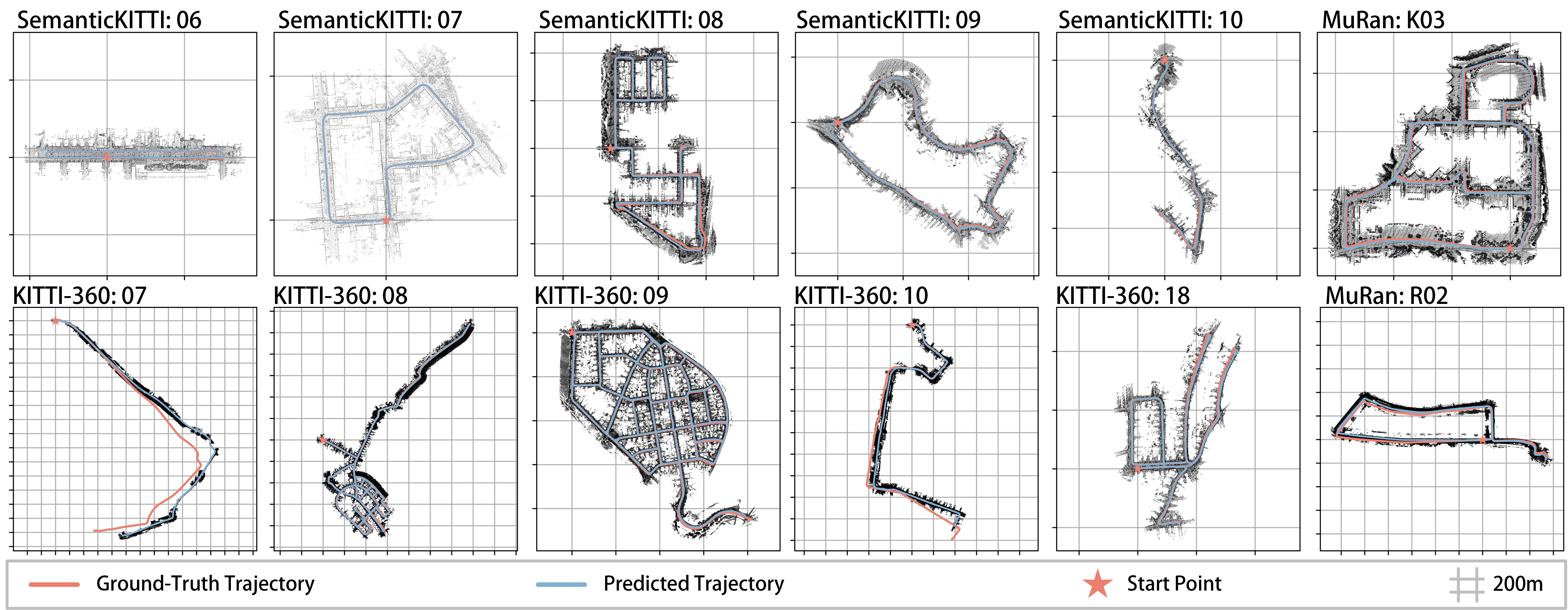}
	\caption{Trajectory estimation result on different datasets.The blue line represents the predicted trajectory; the red line represents the ground-truth trajectory; the red pentagram represents the starting point; one grid represents 200m.
	}
	\label{fig_6}
\end{figure*}
\subsection{Compared with Others}
To validate the performance of ADA-DPM, we conducted the qualitative evaluation on the SemanticKITTI dataset using KITTI's official metrics: the Relative Translation Error (TRANS.$\downarrow$) (\%) and the Average Rotation Error (ROT.$\downarrow$) ($°/100 m$). For some methods, the corresponding data were not provided in their papers, which are indicated by '-' in Table \hyperref[table1]{1}). As shown in Table \hyperref[table1]{1}), ADA-DPM achieved the best performance in four metrics and the second-best performance in two metrics. It can be observed that although ADA-DPM is a LiDAR-based single-modal SLAM method, it still exhibits competitive performance compared to multi-modal methods that utilize both cameras and LiDAR.

Furthermore, to further investigate the environmental bias of ADA-DPM, we conducted comparative experiments using sequences 00–05 from SemanticKITTI, sequences 00, 01, and 03–06 from KITTI-360, as well as six sequences collected in Sejong and DCC from MulRan as the training set. For the test set, we used sequences 06–10 from SemanticKITTI, sequences 07–10 and 18 from KITTI-360, and sequences KAIST 03 (K3) and Riverside 02 (R2) from MulRan. We adopted the Mean Absolute Pose Error (APE$\downarrow$) as the evaluation metric, and compared our results with seven state-of-the-art odometry and SLAM methods. 

\textbf{Trajectory Estimation:} As shown in Figure \hyperref[fig_6]{6}, ADA-DPM achieves high-precision and globally consistent map reconstruction on all selected datasets. In the figure, the red line represents the ground-truth trajectory, the blue line represents the trajectory predicted by ADA-DPM, and the red pentagram marks the starting point of the trajectory. In particular, for long-range localization and mapping tasks like KITTI-360 (where KITTI-360: 09 contains point clouds over 10,000 meters), ADA-DPM still demonstrates favorable performance.
\begin{table*}[ht]
	\caption{Localization accuracy ($APE\downarrow$) on the test sequences. The \textbf{bold} indicates the best result, and the \underline{underline} indicates the second-best result. GeoTransformer is a point cloud registration method, so for GeoTransformer, we perform end-to-end registration to complete the trajectory evaluation.
	}
	\label{table2}
	\centering
	% \LARGE
	\resizebox{\linewidth}{!}{      				
		\begin{tabular}{c|c c c c c   c c c c c  c c}  	
			\toprule
			\midrule
			\multirow{2}{*}{Method} &  	
			\multicolumn{5}{c}{SemanticKITTI} & 
			\multicolumn{5}{c}{KITTI-360} &
			\multicolumn{2}{c}{MulRan} \\ 
			\cmidrule(lr){2-6} \cmidrule(lr){7-11} \cmidrule(lr){12-13} 
			& 06 & 07 & 08 & 09 & 10 & 07 & 08 & 09 & 10 & 18 & K03 & R02 \\ 
			\midrule
			LeGO-LOAM\cite{shan2018lego} & 0.88 & 0.67 & 8.84 & 1.95 & 1.37 & 82.84 & 32.79 & 7.36 & 26.56 & 2.80 & - & - \\
			SC-LeGO-LOAM\cite{kim2021scan} & 1.02 & 1.46 & 6.23 & 8.31 & 1.69 & 47.78 & 8.47 & 22.38 & 9.57 & 6.27 & 3.85 & 28.16 \\
			MULLS\cite{pan2021mulls} & \underline{0.48} & 0.38 & 4.16 & 1.99 & \underline{0.97} & 47.25 & 8.24 & 93.88 & 11.99 & 1.52 & 6.63 & - \\
			CT-ICP\cite{dellenbach2022ct} & 0.56 & 0.43 & 4.07 & \underline{1.29} & \textbf{0.94} & \textbf{14.43} & - & 11.41 & \textbf{7.11} & - & - & - \\
			GeoTransformer\cite{qin2023geotransformer} & 24.38 & 8.71 & 22.68 & 22.14 & 16.36 & - & - & 75.93 & - & 34.79 & 71.95 & - \\
			KISS-ICP\cite{vizzo2023kiss} & 0.61 & 0.35 & 3.58 & 1.32 & 0.94 &  \underline{20.51} & 26.24 & 24.00 &  \underline{8.75} & 2.22 & 12.85 & 27.16 \\
			DeepPointMap\cite{zhang2024deeppointmap} & 0.92 & \textbf{0.26} & \textbf{3.49} & \textbf{1.27} & 1.28 & 93.77 & \underline{5.21} & \underline{1.02} & 10.46 & \textbf{1.28} & \underline{1.73} & \underline{15.67} \\
			\midrule 
			Ours & \textbf{0.46} & \underline{0.31} & \underline{3.79} & \underline{1.29} & 1.50 & 35.73 & \textbf{5.09} & \textbf{1.01} & 11.21 & \underline{1.48} & \textbf{1.62} & \textbf{12.78} \\
			\midrule
			\bottomrule
		\end{tabular}
	}
\end{table*}
\begin{figure}[!t]
	\centering
	\includegraphics[width=3.25in]{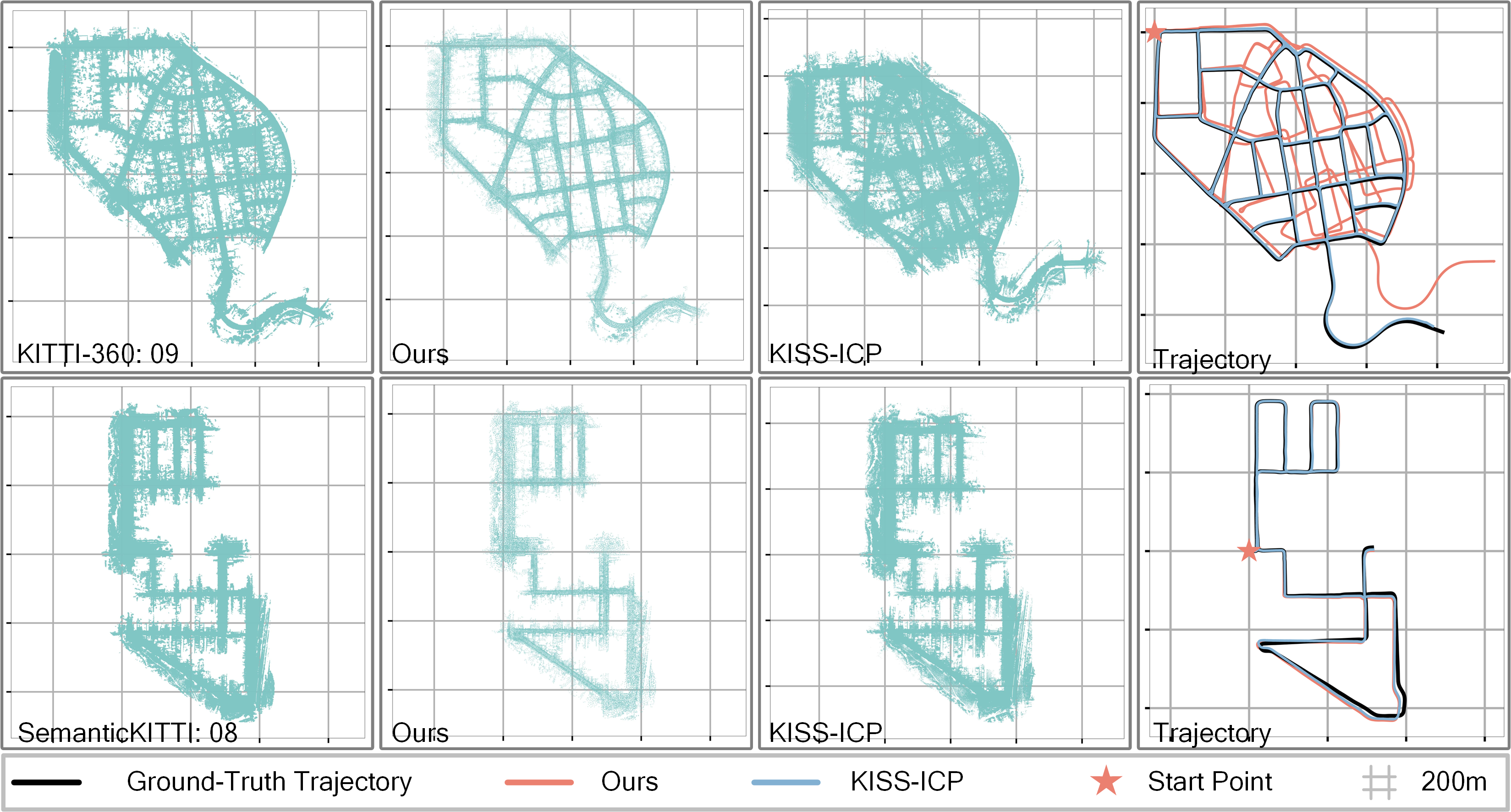}
	\caption{Visualization results of trajectory estimation and map reconstruction compared with the map representation of KISS-ICP.
	}
	\label{fig_7}
\end{figure}
\begin{figure}[!t]
	\centering
	\includegraphics[width=3.25in]{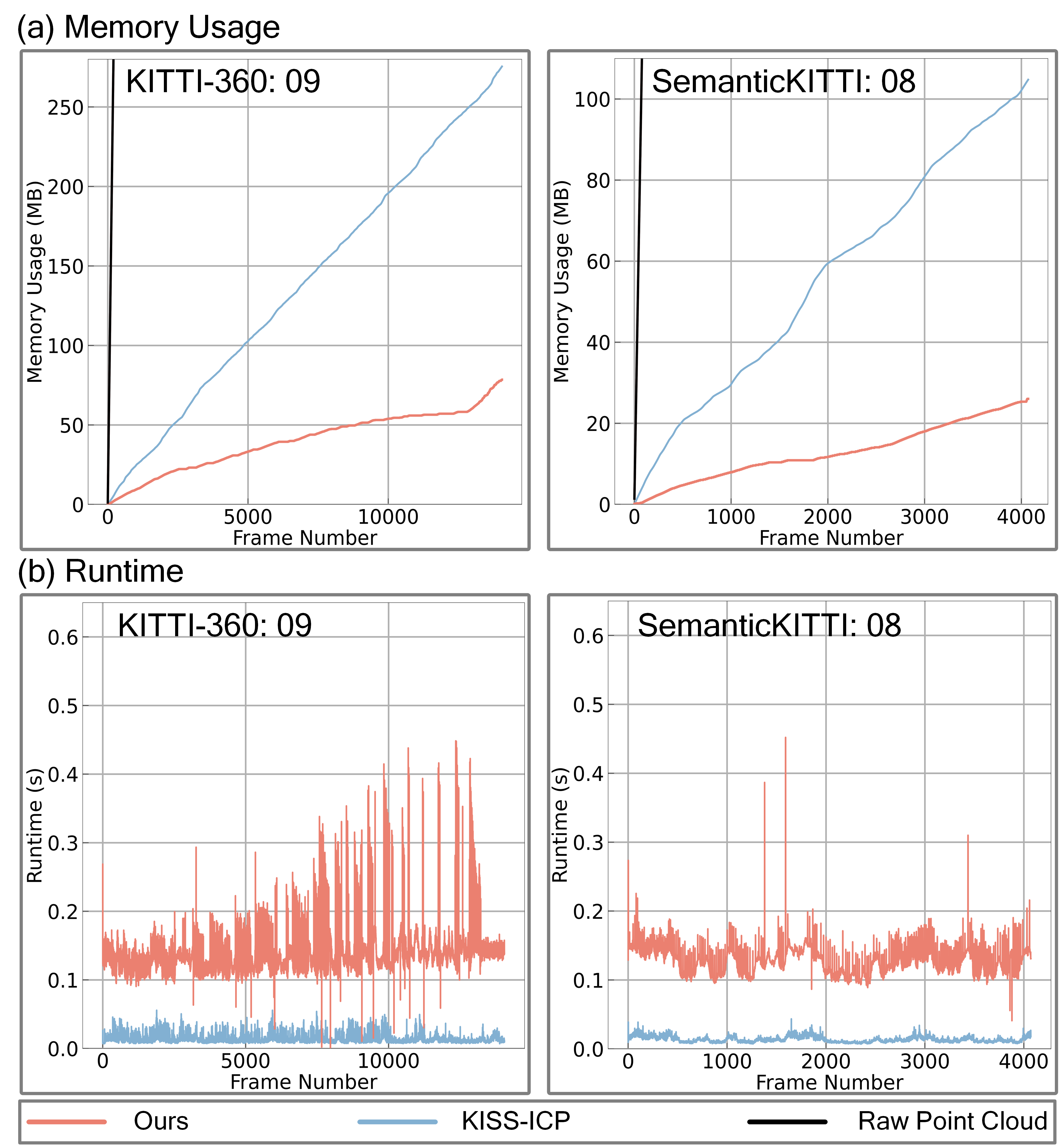}
	\caption{Comparison of memory consumption and runtime for the map representation of KISS-ICP.
	}
	\label{fig_8}
\end{figure}

The trajectory estimation comparison results of ADA-DPM with other algorithms on 12 test sequences are shown in Table \hyperref[table2]{2}. As seen from Table \hyperref[table1]{1}, ADA-DPM achieved the best performance on 5 sequences and the second-best performance on 4 sequences (due to CT-ICP being able to run only based on its own provided data, thus it did not participate in the testing of all sequences).

On the one hand, ADA-DPM benefits from its deep learning-based feature encoding method,which does not rely on distinctive reference objects and has good generalization capabilities for most scenarios (as shown in Table \hyperref[table2]{2}, where traditional methods struggle to generate accurate trajectories for the Riverside02 sequence due to the lack of distinct reference objects). On the other hand, the Dynamic Segmentation Head in ADA-DPM effectively filters out interference from dynamic objects such as vehicles and pedestrians (as demonstrated in Table \hyperref[table2]{2}, where ADA-DPM performs well on urban road sequences in SemanticKITTI with dense vehicles and pedestrians). The Importance Scoring Head further emphasizes feature point pairs contributing to registration (e.g., structural feature point pairs with regular geometric features such as buildings and road signs) while filtering out unstructured feature point pairs (e.g., vegetation with irregular distributions), enabling ADA-DPM to achieve the best performance on the sequences with abundant vegetation (e.g., the Riverside02 and KAIST 03 sequences in Table \hyperref[table2]{2}). Additionally, the GLI-GCN module mitigates the issue of reduced structural similarity in neighborhood structures within overlapping regions caused by a single receptive field, further enhancing localization and mapping accuracy in scenarios with a large number of loop closures (as evidenced by the precise localization and mapping of ADA-DPM on KITTI-360: 09).

\textbf{Map Reconstruction:} To further evaluate the performance of ADA-DPM in map reconstruction task, we visualized the trajectories predicted by ADA-DPM and KISS-ICP, as well as the reconstructed maps, on SemanticKITTI:08 and KITTI-360:09. In Figure \hyperref[fig_7]{7}, the first column shows the ground-truth map, i.e., the global map reconstructed from the original point cloud using ground-truth poses; the second and third columns present the reconstructed maps by ADA-DPM and KISS-ICP, respectively; the last column displays the trajectories predicted by ADA-DPM and KISS-ICP. As shown in the figure, compared to KISS-ICP, ADA-DPM constructs more accurate global maps using less memory and achieves higher localization accuracy. Especially for long-range sequences like KITTI-360:09, ADA-DPM realizes more precise globally consistent localization and mapping compared to KISS-ICP.
\begin{figure}[!t]
	\centering
	\includegraphics[width=3.25in]{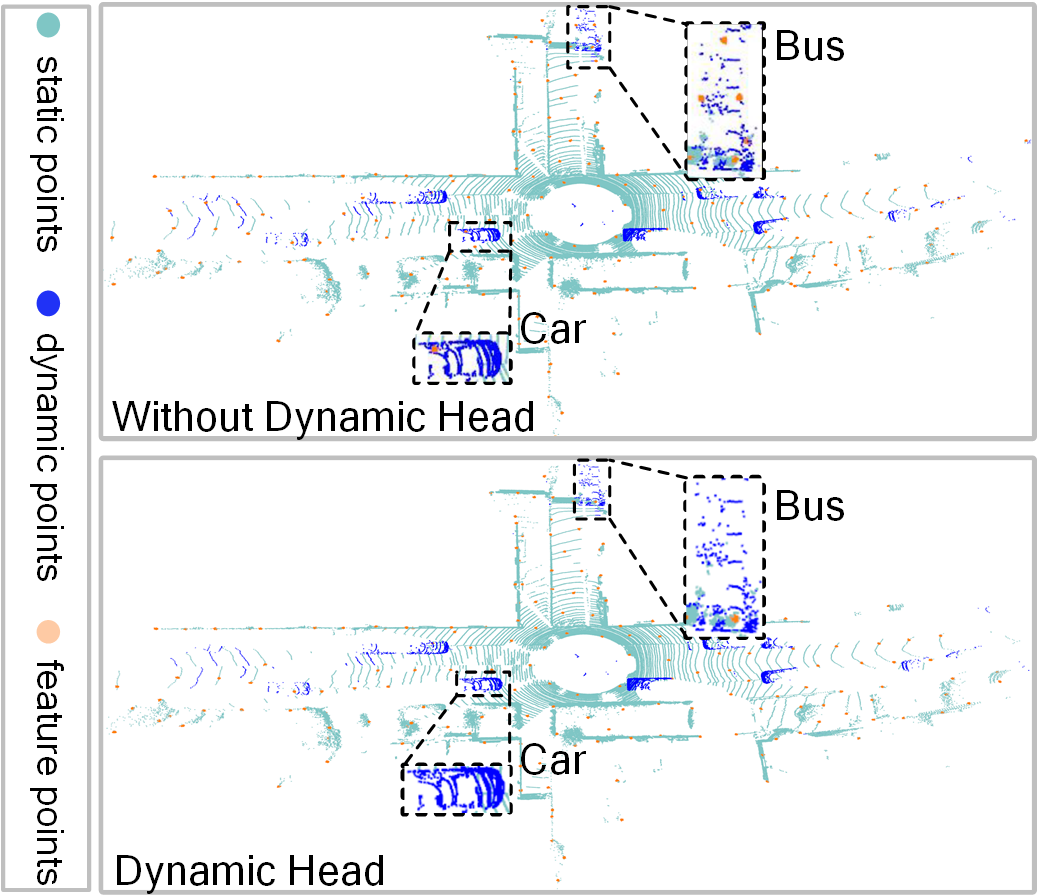}
	\caption{Visualization of feature point sampling.
	}
	\label{fig_9}
\end{figure}
\begin{table}[!t]
	\caption{Trajectory estimation results ($APE\downarrow$) on the validation sequences.The \textbf{bold} indicates the best result}
	\label{table3}
	\centering
	\resizebox{\linewidth}{!}{      				
		\begin{tabular}{c|c c c | c c c }  	
			\toprule
			\midrule
			\multirow{2}{*}{Method} &  	
			\multirow{2}{*}{Dynamic Head} & 
			\multirow{2}{*}{Scoring Head}&
			\multirow{2}{*}{GLI-GCN} &
			\multicolumn{3}{c}{Without Noise $\setminus$ Noise} \\ 
			\cmidrule(lr){5-7} 
			& & & & 00 & 01 & 02 \\ 
			\midrule
			A &  &  & & 2.26$\setminus$3.29 & 7.96$\setminus$10.21 & 6.02$\setminus$7.01\\
			B & $\surd$ & & & 2.01$\setminus$2.72 & 8.02$\setminus$10.82 & 5.21$\setminus$7.44\\
			C & $\surd$ & $\surd$ &  & 2.64$\setminus$2.84 & \textbf{5.16}$\setminus$5.62& 5.56$\setminus$6.94 \\
			D & $\surd$ & $\surd$ & $\surd$ & \textbf{0.91}$\setminus$\textbf{1.67} & 5.26$\setminus$\textbf{5.28} & \textbf{3.26}$\setminus$\textbf{4.32} \\
			\midrule
			\bottomrule
		\end{tabular}
	}
\end{table}

Furthermore, to evaluate the map representation efficiency of ADA-DPM during large-scale map reconstruction, we compared the localized storage data sizes of ADA-DPM and KISS-ICP on two large-scale sequences (SemanticKITTI: 08 and KITTI-360: 09; all data are stored in float32 data type). The results are shown in Figure \hyperref[fig_8]{8(a)}. Benefiting from ADA-DPM's adoption of deep learning-based sparse point cloud feature descriptors (we used the same map representation as DeepPointMap), ADA-DPM reduced memory usage by 60\% compared to KISS-ICP on SemanticKITTI:08 and approximately 50\% on KITTI-360:09. The runtime of ADA-DPM and KISS-ICP are shown in Figure \hyperref[fig_8]{8(b)}. ADA-DPM processes a single frame in approximately $0.15–0.5$~s, while KISS-ICP processes a frame in $0.01–0.05$~s. Compared with KISS-ICP, ADA-DPM has no advantage in running speed. However, ADA-DPM performs better in global map reconstruction and trajectory estimation, and it also has a significant advantage in map memory usage.
\begin{table}[!t]
	\caption{Trajectory estimation results ($APE\downarrow$) on the validation sequences under different noise ratios.}
	\label{table4}
	\centering
	\begin{tabularx}{\columnwidth}{l X X X p{2cm}}   % Last column as fixed-width (2cm)
		\toprule
		\midrule
		\multirow{2}{*}{\textbf{Noise Ratio}} & \multicolumn{4}{c}{\textbf{SemanticKITTI}} \\
		\cmidrule(lr){2-5}
		& \textbf{00} & \textbf{01} & \textbf{02} & \textbf{Mean} \\ 
		\midrule
		0 \% & 0.91 & 5.26 & 3.26  & 3.14 \\
		0$-$10 \% & 1.67 & 5.28 & 4.32  & 3.76 \textcolor{red}{$+19\%$} \\
		0$-$20 \% & 2.03 & 5.75 & 4.04  & 3.94 \textcolor{red}{$+25\%$} \\
		0$-$30 \% & 2.16 & 5.98 & 4.53  & 4.22 \textcolor{red}{$+34\%$} \\
		0$-$40 \% & 1.72 & 5.79 & 4.80  & 4.10 \textcolor{red}{$+30\%$} \\
		0$-$50 \% & 2.26 & 5.63 & 4.64  & 4.18 \textcolor{red}{$+33\%$} \\
		\midrule
		\bottomrule
	\end{tabularx}
\end{table}
\begin{figure}[!t]
	\centering
	\includegraphics[width=3.25in]{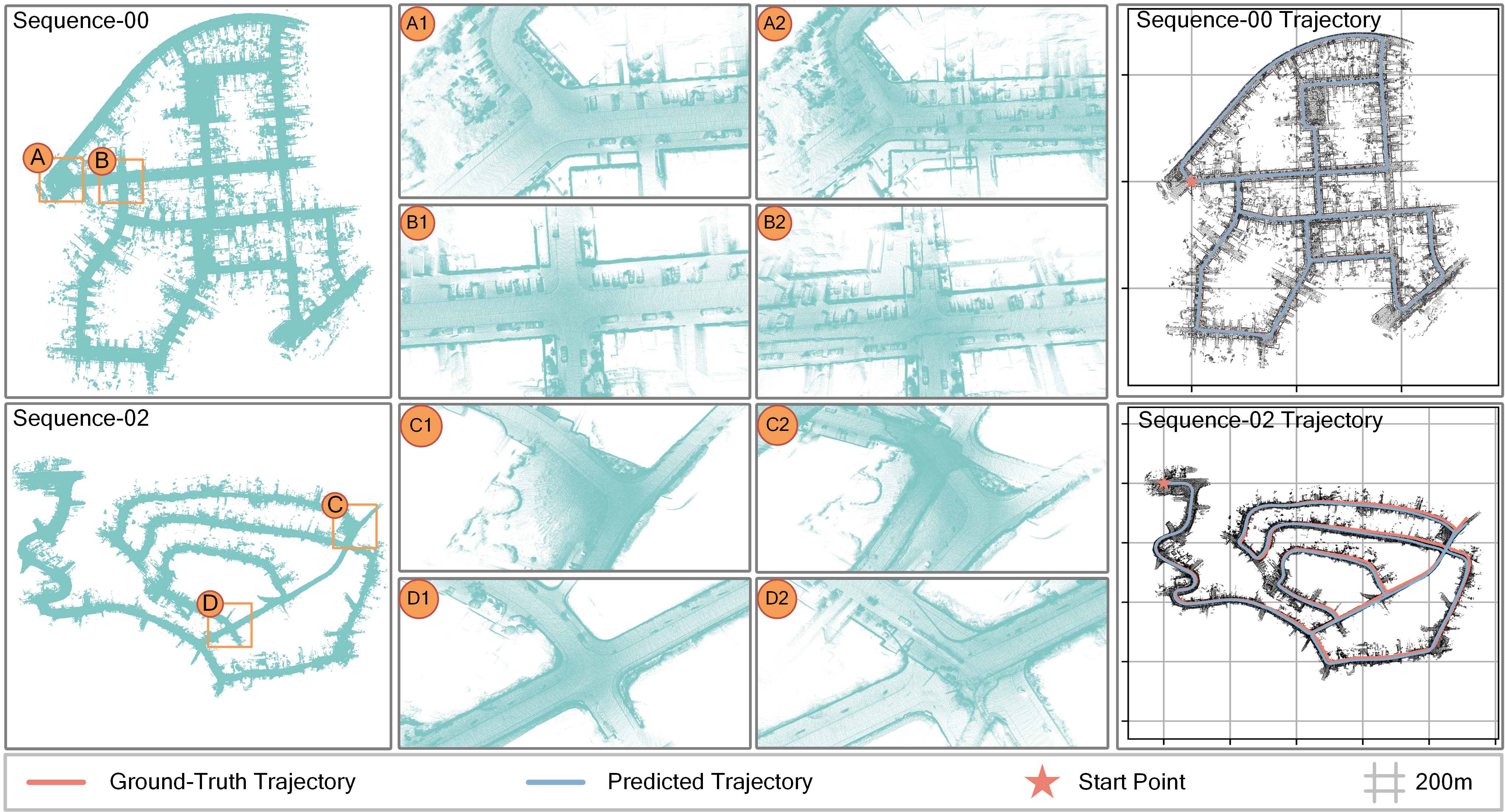}
	\caption{Global consistent dense mapping results on SemanticKITTI Sequence00, 02.The blue line represents the predicted trajectory; the red line represents the ground-truth trajectory; the red pentagram represents the starting point.
	}
	\label{fig_10}
\end{figure}
\subsection{Ablation experiments}
To further validate the effectiveness of each module in ADA-DPM, we conducted ablation experiments. All ablation experiments are performed on the same dataset partitioning (sequences 03-10 of SemanticKITTI are selected as training sequences, and sequences 00-02 are used as validation sequences).

As shown in Table \hyperref[table3]{3}, we separately removed the GLI-GCN module, Importance Scoring Head, and Dynamic Segmentation Head to assess their importance. To validate the robustness of the Importance Scoring Head to noise points, we compared test results under noise and noise-free conditions. We simulated noise conditions by randomly adding noise points sampled from the normal distribution $\mathcal{N}(0, 0.1^2)$ to 0\%–10\% of the point cloud, with noise clipped to the range $[-0.2, 0.2]m$ on each axis. In Table \hyperref[table3]{3}: Method D refers to ADA-DPM; Method C is ADA-DPM with the GLI-GCN module removed; Method B is Method C with the Importance Scoring Head removed; Method A is Method B with the Dynamic Segmentation Head removed.

Compared Method A with Method B, it can be observed that for urban scenes such as sequences 00 and 02, which contain many dynamic objects (e.g., cars and pedestrians), adding the Dynamic Segmentation Head leads to a decrease in APE. We visualized the feature points of Method A and Method B to qualitatively analyze the effect of the Dynamic Segmentation Head. As shown in Figure \hyperref[fig_9]{9}, after adding the Dynamic Segmentation Head, only a small number of the feature points sampled by Method B correspond to dynamic objects, demonstrating that the Dynamic Segmentation Head can partially filter out the interference from dynamic objects.

When compared Method B with Method C, although the APE on sequences 00 and 02 slightly increases after incorporating the Importance Scoring Head, under conditions with added random noise, Method C exhibits a significantly lower APE than Method B on sequence 01, while the difference between the two methods is minimal on sequences 00 and 02. The reason may lie in the fact that for scenes with frequent loop closures (e.g., sequences 00 and 02), loop closure detection can mitigate cumulative errors caused by noise points to some extent; however, for scenes with fewer loop closures (e.g., sequence 01), it is challenging to eliminate cumulative errors through loop closure detection alone. Method C, by leveraging the Importance Scoring Head to assign contribution scores to each feature point pair, can mitigate the impact of noise points to a certain degree, with more pronounced effects particularly in scenes with fewer loop closures. Finally, compared Method D with Method C, although Method D shows a slight increase in APE on sequence 01, the added GLI-GCN module further enhances the mapping accuracy in loop-closure regions for sequences 00 and 01.

Furthermore, to further validate the robustness of ADA-DPM under noise conditions, we compared test results under different noise point ratios (with maximum noise point ratios of 10\%, 20\%, 30\%, 40\%, and 50\%), as shown in Table \hyperref[table4]{4}. As can be observed from Table \hyperref[table4]{4}, even when noise is applied to 50\% of the point cloud, ADA-DPM is still capable of completing globally consistent trajectory estimation.

Finally, to further validate the effectiveness of the GLI-GCN module in regions with loop closures, we visualize the mapping results of the full ADA-DPM and the ADA-DPM with the GLI-GCN module removed, as shown in Figure \hyperref[fig_10]{10}. In Figure \hyperref[fig_10]{10}, positions A, B, C, and D represent four loop-closure locations, where A1, B1, C1, and D1 denote the mapping results of ADA-DPM, and A2, B2, C2, and D2 denote the mapping results of ADA-DPM without GLI-GCN module. As shown in Figure \hyperref[fig_10]{10}, the maps constructed by ADA-DPM at each loop-closure position exhibit consistent alignment. This successful alignment can be attributed to GLI-GCN, which utilizes different neighborhoods to perceive receptive fields of varying sizes. By fusing and emphasizing shallow features that have high correlation with deep feature points, it generates local feature descriptors with richer geometric characteristics. This alleviates the issue of reduced neighborhood structure similarity in overlapping regions caused by a single receptive field, thereby further enhancing localization and mapping accuracy in the overlapping areas.
\section{Conclusions}
In this paper, to address the challenges of dynamic objects, noise interference, and reduced similarity in loop-closure regions in LiDAR SLAM for autonomous mobile robots, we propose a neural descriptors-based adaptive noise filtering strategy for SLAM, named ADA-DPM. (1) We construct the Dynamic Segmentation Head to predict the probability of a feature point belonging to a dynamic object. By filtering out dynamic points, Dynamic Segmentation Head eliminates the interference of ego-motion from dynamic objects on pose estimation. (2) We design the Global Importance Scoring Head to assign global contribution scores to feature point pairs. It emphasizes the contribution of structural points to achieve precise registration while discarding noise and unstructured points. Additionally, we introduce the importance score-weighted global reconstruction loss function. It not only optimizes registration accuracy but also avoids trivial solutions where the network generates all-zero scores. More importantly, it imposes implicit supervision, encouraging the network to learn feature points with greater contribution to point cloud registration tasks. (3) We propose the GLI-GCN module to construct cross-layer neighborhood graphs. It alleviates the issue of reduced similarity in overlapping regions caused by low-resolution features and captures richer geometric features, thereby enhancing mapping accuracy in loop-closure regions. Experimental validations on multiple datasets demonstrate that ADA-DPM significantly enhances robustness to dynamic objects and noise, outperforming state-of-the-art methods in challenging scenarios. Furthermore, GLI-GCN is shown to further improve mapping accuracy in scenarios with frequent loop closures.

Although ADA-DPM achieves favorable performance in both location accuracy and system robustness, compared to traditional methods, deep learning-based methods require more precise annotations, such as accurate robot poses. However, in real-world scenarios, SLAM label quality is often insufficient, which may affect experimental results. Additionally, our method performs relatively poorly in certain remote rural scenes (e.g., KITTI-360: 07 and 10), likely due to sparse reference objects and limited geometric information in the point clouds. In the future, we plan to adopt the strategy that combines vision and LiDAR, using vision-based feature encoding to perceive semantic information in the scene to compensate for the lack of geometric information.

Furthermore, compared with KISS-ICP, ADA-DPM has no advantage in terms of speed. On one hand, deep learning-based methods generally require more GPU computational power. In future work, PointNeXt could be replaced with a more lightweight network.
\bibliographystyle{IEEEtran}
\bibliography{reference.bib}	
	
\end{document}